\pdfoutput=1

\documentclass[11pt,authoryear,round]{article}
\usepackage{siunitx}
\usepackage{float}
\usepackage{booktabs}  
\usepackage{xcolor}

\usepackage[final]{acl}

\usepackage{times}
\usepackage{latexsym}
\usepackage{enumitem}
\usepackage{booktabs} 
\usepackage{siunitx} 
\usepackage{tabularx} 
\usepackage{CJKutf8}
\newcommand{\zh}[1]{\begin{CJK*}{UTF8}{gbsn}#1\end{CJK*}}
\usepackage[T1]{fontenc}

\usepackage[utf8]{inputenc}

\usepackage{microtype}

\usepackage{inconsolata}

\usepackage{graphicx}

\title{CAPC-CG: A Large-Scale, Expert-Directed LLM-Annotated Corpus of Adaptive Policy Communication in China}

\author{
 \textbf{Bolun Sun\textsuperscript{1,2}},
 \textbf{Charles Chang\textsuperscript{3}},
 \textbf{Yuen Yuen Ang\textsuperscript{1}},
 \textbf{Ruotong Mu\textsuperscript{1}},
 \\
 \textbf{Yuchen Xu\textsuperscript{1}},
 \textbf{Zhengxin Zhang\textsuperscript{1}},
 \textbf{Pingxu Hao\textsuperscript{1}}
\\
\\
 \textsuperscript{1}Johns Hopkins University,
 \textsuperscript{2}Northwestern University,
 \textsuperscript{3}Duke Kunshan University
\\
}

\begin{document}
\maketitle
\begin{abstract}
We introduce \textbf{CAPC-CG}, the \textit{Chinese Adaptive Policy Communication (Central Government) Corpus}, the first open dataset of Chinese policy directives annotated with a five-color typology of policy signals, capturing clarity and ambiguity, grounded in the theory of adaptive policy communication. Spanning 1949–2023, this corpus includes laws, regulations, and rules issued by Chinese central authorities, segmented into 3.3 million paragraph units. We further propose and validate an expert-directed LLM annotation method that integrates codebook design, structured training, a two-step workflow, and LLM-based scaling. Alongside the corpus, we release metadata and a gold-standard labeled set developed by trained coders. Inter-annotator agreement achieves a Fleiss' kappa of \(\kappa=0.86\) on directive labels, indicating high reliability. We provide baseline classification results with several large language models (LLMs), together with our codebook, and describe patterns from the data. This release enables downstream tasks and multilingual NLP research in communication strategies under complexity and uncertainty. 

\end{abstract}

\section{Introduction}

How do central authorities steer policy implementation under conditions of complexity and uncertainty? \textit{Adaptive policy communication} is a theory of governance in such settings, where leaders exercise influence rather than precise control by combining clear and ambiguous instructions to calibrate discipline and flexibility \cite{ang2016china}. This logic applies broadly to large, decentralized organizations, including the communication of complex issues such as geopolitics or central bank policy \cite{blinder2008central}, but is especially salient in China’s governing model of "directed improvisation."

China's one-party political system is politically centralized yet administratively and economically decentralized \citep{landry2008decentralized, lieberthal&okenberg1988policymaking, ang2016china}. The Communist Party and the State Council communicate instructions through written policy directives, which local government actors then interpret and implement. These directives—including National Laws, Administrative Regulations, and Ministerial Rules—function like the nervous system, sending signals from central to local authorities across policy domains and regions. 

As \citet{ang2016china, ang2023ambiguity} demonstrates, these directives operate along a spectrum of clarity and ambiguity. Some directives clearly authorize or forbid, while others deliberately use ambiguous language to allow local flexibility and adaptation. Operationalizing this adaptive logic at scale requires systematic annotation, yet no dataset has previously existed. We therefore present the first such resource: \textbf{the Chinese Adaptive Policy Communication Corpus (CAPC-CG)},\footnote{\href{https://github.com/Baron-Sun/CAPC-CG}{Dataset and Scripts}}
an open corpus of central-government policy directives spanning 1949–2023.

Inspired by expert-annotated corpus construction such as \textsc{OPP-115} \cite{wilson-etal-2016-creation}, we design an \textit{expert-directed LLM annotation method} that integrates structured annotation with scalable modeling. This method combines an expert-designed codebook, rigorous annotator training, a two-round labeling workflow, and LLM fine-tuning to construct and scale high-quality labels. 

Building on \citeauthor{ang2023ambiguity}'s original typology of three policy signals (\textit{Black = Authorizing}, \textit{Red = Prohibiting}, \textit{Grey = Ambiguous}), we extend it into five categories by adding \textit{Yellow (Pressuring)} and \textit{Charcoal (Flexible)}. This expanded typology captures finer variation in policy signals and, in particular, it captures \textit{Yellow} and \textit{Charcoal} signals that have become more salient under Xi Jinping's leadership. 

Using this method, we construct CAPC-CG, released with code, guidelines, and baseline models to support reproducible NLP research. Drawing on expert knowledge of the Chinese bureaucracy, we design a detailed codebook that operationalizes Chinese policy language into codable categories. 
We trained Chinese-proficient researchers to annotate a gold-standard dataset and fine-tuned large language models (LLMs) to scale annotation across the corpus. Inter-annotator agreement reaches \(\kappa=0.86\), indicating highly reliable labels despite the heterogeneity of policy language \cite{bhowmick2008agreement,richie2022inter}. 

We highlight three contributions:
\begin{enumerate}
  \item We operationalize a five-color typology of policy signals, grounded in a theory of adaptive policymaking, yielding a gold-standard labeled set of about 6,000 paragraphs with high inter-annotator agreement.
  \item  We produce a large‐scale, computable corpus of Chinese central-government policy directives, with normalized metadata, de-duplication, and paragraph-level segmentation, addressing data scarcity in Chinese policy corpora.
  \item We propose and validate an expert-directed annotation method, providing evaluation splits, scripts, and baseline models using LLM.\end{enumerate}

All data, code, and guidelines are released under a permissive license, establishing \textbf{CAPC-CG} as a standardized testbed for research on policy communication, including diachronic analysis, cross-regime comparisons, information extraction, compliance monitoring, and policy-impact assessment.

\section{Theoretical Foundation: Adaptive Policy Communication}

Although several large-scale Chinese datasets have been developed, they are mainly based on Internet and social media content, often with limited quality control \citep{wang2025opencsg, li2024mapcc, zhang2023chinesewebtext, yuan2023wanjuan, yuan2021wudaocorpora, xu2020cluecorpus2020}. In contrast, resources for political and legal documents remain scarce: existing datasets lack high-quality source texts and rigorous annotations, limiting the ability to trace policymaking at the national level \citep{ahrens2018using, xiao2018cail2018, yao2022leven, duan2019cjrc, wang2025improving, li2025topic, xiao2021lawformer}. This gap not only limits research in computational linguistics, but also social science research on lawmaking and policymaking in China and comparatively. 

Among scholars of Chinese bureaucracy, the long-standing impression has been that ``the higher levels give vague directives'' \citep[p.~340]{lieberthal&okenberg1988policymaking} or pronounce ``catchy slogans that... direct the flow of policy implementation at all levels'' \citep[p.~16]{brandt2008china}. But this conventional view raises a puzzle: How can a vast country like China, facing immense challenges and numerous tasks, be effectively governed through slogans and vague expressions from the top? 

\citet{ang2016china} advanced a revisionist account: adaptive policy communication. In \textit{How China Escaped the Poverty Trap}, she characterizes post-1979 Chinese governance as a system of ``directed improvisation,'' in which central authorities signal and steer rather than dictate, while local governments improvise, producing diverse paths of policy implementation and development. A key mechanism of ``directing'' is that central authorities apply varying mixtures of ambiguous and clear mandates—declaring and enforcing guardrails while allowing flexibility in selected domains or time periods. 

More precisely, Ang highlights three varieties of directives: \textit{Black} (clearly states what can be done), \textit{Red} (clearly states what cannot be done), and \textit{Grey} (ambiguous about what can or cannot be done). In other words,\textit{ Black} authorizes,\textit{ Red }prohibits, while \textit{Grey} implicitly permits flexibility. Previously, \citet{ang2023ambiguity} operationalized these three categories using a pilot sample of central regulations. In this project, we refine classification and annotation methods using a hybrid of LLMs and trained human coders, and scale the analysis to a much larger dataset. 

To achieve precision, we segment the vast corpus of highly variegated and lengthy bureaucratic texts into smaller, analyzable paragraphs, producing gold-standard, expert-labeled data for NLP research. We also share a rigorous and replicable workflow that combines source data and human annotation, helping mitigate bias and inaccuracy in LLM applications \cite{baumann2025}. 

\section{Corpus Creation and Structure}
\begin{figure}[htbp]
    \centering
    \includegraphics[width=1\linewidth]{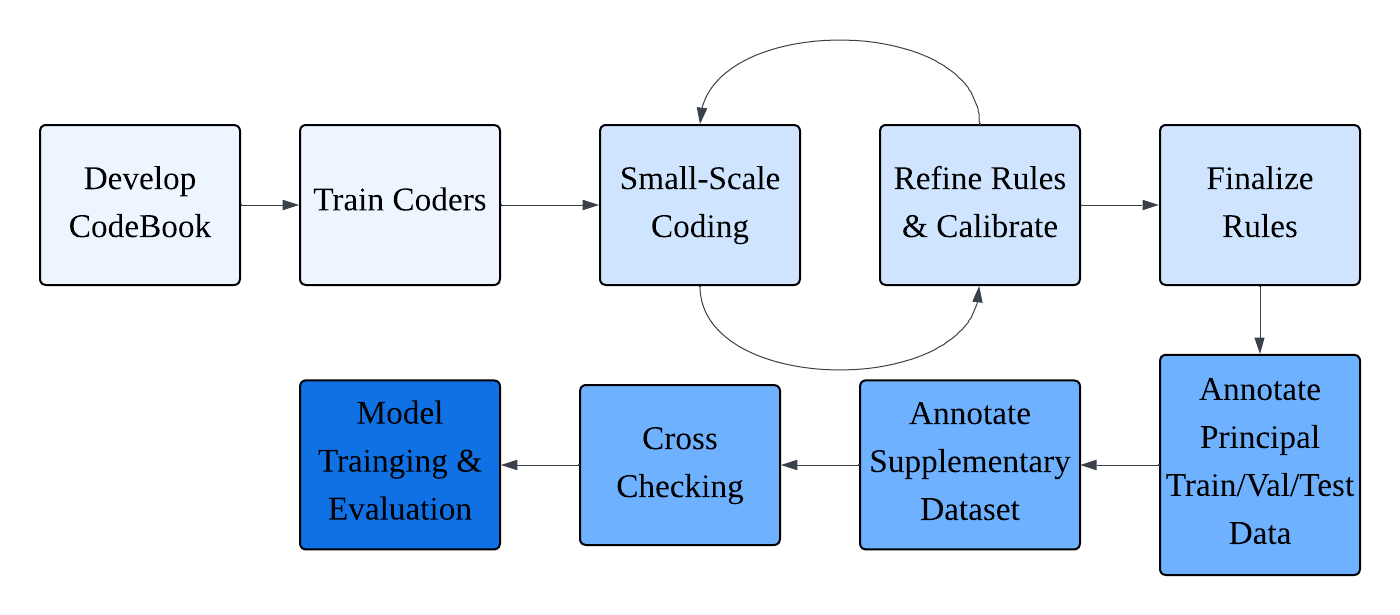}
    \caption{Workflow for Corpus Creation}
    \label{fig: Workflow for Corpus Collection and Annotation}
\end{figure}

This section details how we created \textbf{CAPC-CG}, from our document collection, annotation schema, to the resulting metadata-rich dataset. Figure \ref{fig: Workflow for Corpus Collection and Annotation} illustrates our workflow step by step.

\subsection{Scope of Policy Directives}
\label{subsec:policy_selection}

Our analysis focuses on policy directives, defined as formally issued documents in which a higher-level authority addresses state or public-sector actors with actionable
 implications—that is, the bureaucrats are expected to notice, interpret, and respond to the directives. Such directives form the ``vertical formal channels'' of bureaucratic communication, as \citet{oksenberg1974methods} noted. Central directives define national priorities and directions, then as they travel down the hierarchy, each level of government further interprets and details their implementation of the instructions. 

To construct our data set, we collected three types of central directives. 1)\textbf{National Laws} and decisions promulgated by the National People's Congress (NPC), the legislative branch. Among the directives, these laws carry the highest level of authority. 2) \textbf{Administrative Regulations}, which are typically issued by the State Council (the highest office of the executive branch), or the General Office of the Communist Party Central Committee, or are co-issued by the State Council, the Communist Party Central, or ministries. 3) \textbf{Ministerial Rules} issued by central-level ministries and commissions.

This dataset spans 1 October 1949 to 13 March 2023, from the founding of the People's Republic of China through the end of Xi Jinping's second term. It covers all the central leaders since founding, from Mao Zedong, Deng Xiaoping, Jiang Zemin, Hu Jintao, to Xi Jinping.

\begin{table}[ht]
    \centering
    \begin{tabular}{p{5cm}|p{1.7cm}}
        \hline
        Year & 1949-2023 \\
        Policy Documents & 337,038 \\
        Policy Paragraphs & 3,275,477 \\
        \hline\hline
        Annotated Policy Paragraphs & 2,647,695 \\
        Mean TextLength & 162.50 \\
        Median TextLength & 108.00 \\
        \hline
\end{tabular}

\vspace{0.2em}
{\footnotesize \textit{Note:} ``Annotated Policy Paragraphs'' includes all labeled paragraphs, including Neutral (N), and excludes segments shorter than 20 Chinese characters as well as formatting noise.}
\caption{Summary Statistics}
\label{tab:statistics}
\end{table}

For each document, we parsed and created a standardized metadata record with the following fields: \textit{title}, \textit{validity status}, \textit{hierarchical level of legal force}, \textit{issuing body}, \textit{reference number}, \textit{date of promulgation}, and \textit{date of entry into force}. The product is a massive longitudinal corpus. Spanning multiple leadership eras, it contains 337,038 unique policy documents. Table~\ref{tab:statistics} presents summary statistics of the corpus, including its scale before and after annotation at the paragraph level.

Next, we transformed the raw text into a structure that captures our five-category schema: \textit{Black (Authorizing), Red (Prohibiting), Grey (Ambiguous), Yellow (Pressuring), Charcoal (Flexible)}, as presented in Table~\ref{tab:signals}.\footnote{Colors are employed solely as mnemonic devices; the analytical model encodes the abstract categories themselves.} Appendix~\ref{app:codebook} provides our codebook that details the classification rules. 

\begin{table*}[t]
  \centering
  \renewcommand{\arraystretch}{1.5}
  \small
  \begin{tabularx}{\textwidth}{@{}l p{0.2\textwidth} p{0.2\textwidth}@{} p{0.44\textwidth}}
    \toprule
    \textbf{Colour} & \textbf{Governmental Intent} & \textbf{Signal to Local Actors} & \textbf{Example}\\
    \midrule
    Red (R) & Prohibiting & ``No, you \emph{cannot}.'' & No agency shall intercept or misappropriate the subsidy funds for any reason or in any form.\\
    Grey (G) & Ambiguous & ``You \emph{may} or \emph{may not}.'' & This matter will be left to ``relevant departments'' to weigh the pros and cons and decide for themselves. \\
    Charcoal (C) & Flexible & ``Yes, do it \emph{flexibly}.'' & Actively promote this technology through policy experiments and tailoring to local contexts. \\
    Black (B) & Authorizing & ``Yes, you \emph{can}.'' & The government encourages, supports, and steers the development of the non-public sector. \\
    Yellow (Y) & Pressuring & ``Yes, you \emph{must}.'' & This target must be achieved by any means necessary and will determine bureaucratic performance. \\
    \bottomrule
  \end{tabularx}
  \caption{Five-Color Typology of Policy Signals}
  \label{tab:signals}
\end{table*}

We segmented all documents into paragraphs using an ensemble of rule-based algorithms and LLM (see the detailed segmentation method in Appendix~\ref{sec:segmentation}). Because both the structure and language of Chinese policy directives are highly heterogeneous, a single document can contain multiple paragraphs and policy signals. To tackle this challenging degree of variation, we apply an LLM to segment each document into units bounded by a coherent signal \citep{merola2025}. We assigned annotators to manually validate 4,600 randomly sampled segmented paragraphs and found that 97.74\% were correctly segmented. Following this validation, we retained non-actionable units under the \textit{Neutral} (N) label (see Appendix~\ref{app:codebook}) while filtering segments shorter than 20 Chinese characters and formatting noise. This procedure produced a final set of 2,647,695 paragraphs.

\subsection{Annotation Scheme and Process}
\label{subsec:annotation}

\begin{figure}[htbp]
    \includegraphics[width=0.5\textwidth]{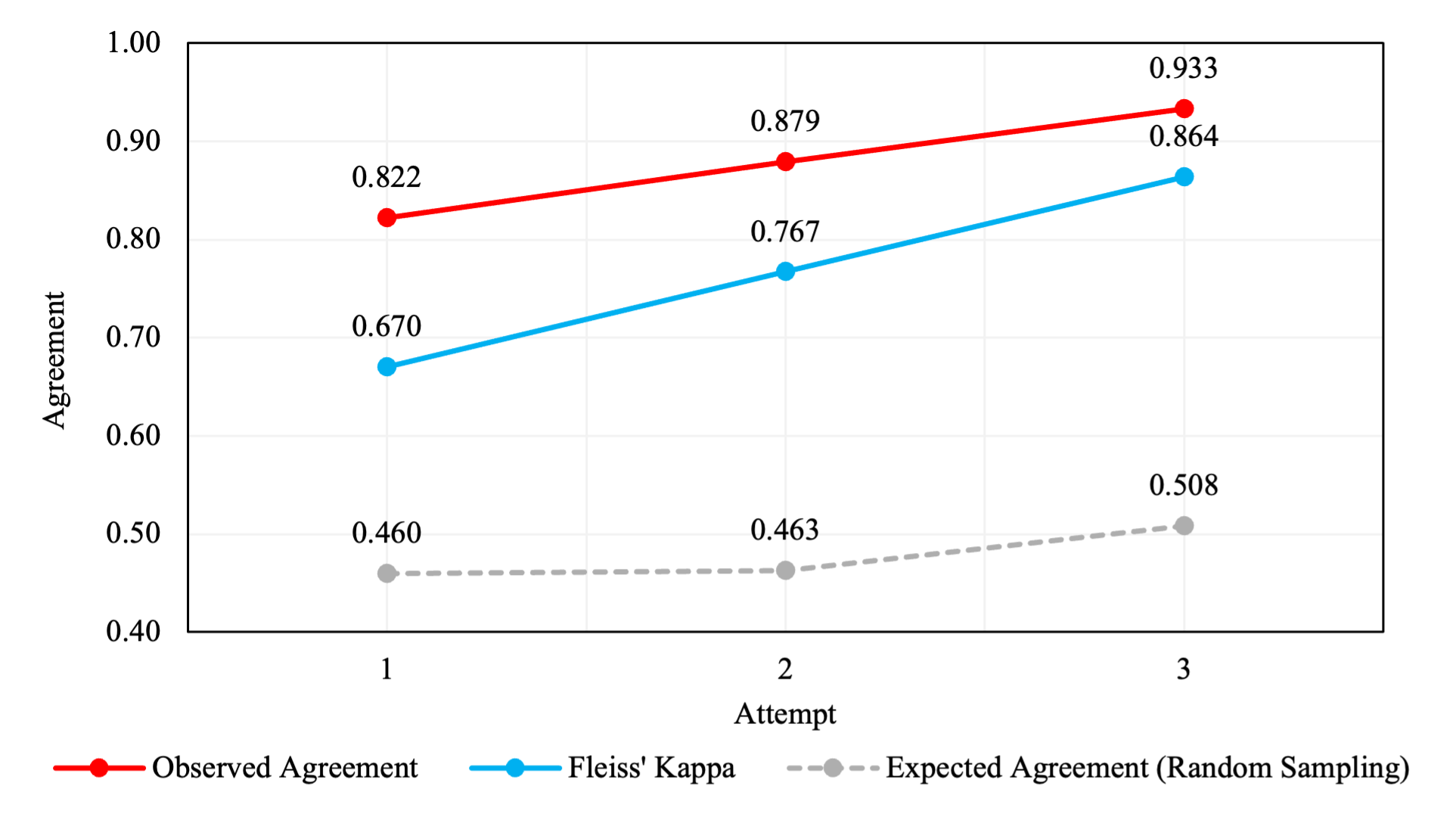}
    \footnotesize \textit{Note:} Expected Agreement fluctuation (0.460 $\rightarrow$ 0.508) represents natural variance in label marginal distributions across small non-identical batches (n=500).
    \caption{Inter-Coder Agreement Improvement}
    \label{fig:kappa}
\end{figure}

We created a gold-standard dataset through a rigorous, multi-stage annotation procedure carried out by three domain experts \citep{marcinkiewicz1994building, tseng2020best}, who annotated paragraph-level texts over a three-month training period. As preparation, all three coders read a diverse sample of policy directives carefully and analyzed representative cases of the five categories listed in Table ~\ref{tab:signals}. Then, we organized a workshop dedicated to explaining and testing the codebook. Every week, the annotators coded small samples of roughly 500 paragraphs, compared their annotation results, then discussed the results and refined their interpretation of the coding rules. By the end of the training period, our annotators reached $K=0.864$ of inter-coder reliability. Figure~\ref{fig:kappa} shows the inter-coder agreement over three rounds of independent coding in the training phase, with no discussion among annotators during the coding process. While the expected agreement under random sampling remained largely unchanged, Fleiss' Kappa improved substantially from 0.670 to 0.864 after iterative training and calibration. The improvement resulted from strengthened coding rules after iterative training, not from a forced consensus, indicating that the three annotators reached high agreement and the dataset is robust. 

Once we were confident about the annotators' grasp of the codebook, they proceeded to code paragraphs at scale. Each week, we assigned 1,000 randomly sampled paragraphs and repeated this process without sample replacement. By the end of the period, they annotated 6,000 samples, ensuring an even distribution across all labels.

\begin{figure}[htbp]
    \centering
    \includegraphics[width=1\linewidth]{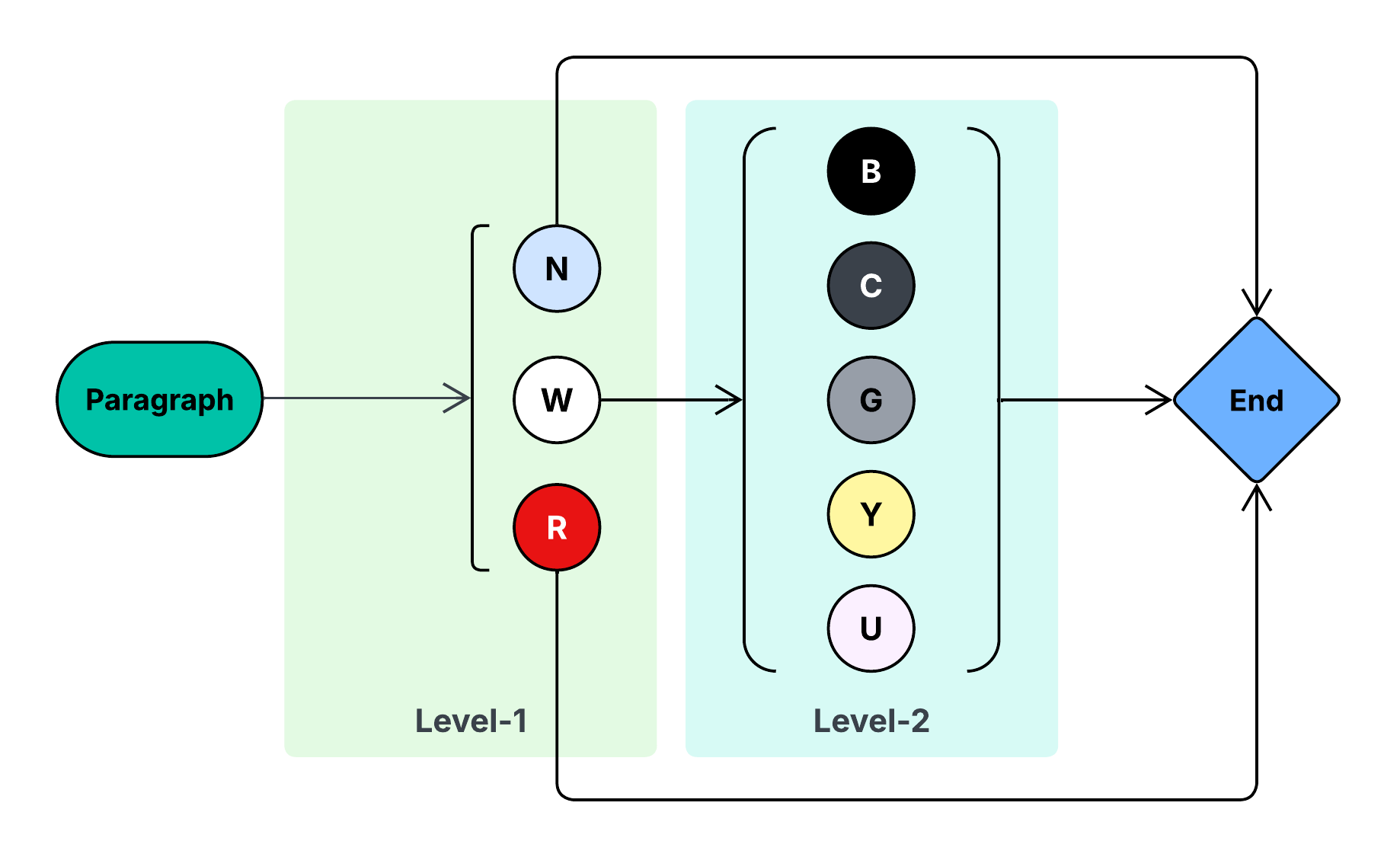}
    \caption{Two-Round Labeling Workflow}
    \label{fig:document_layer_pie}
\end{figure}

\paragraph{Two-Round Labeling Workflow}
Each paragraph is processed through a two-step classification procedure (Figure ~\ref{fig:document_layer_pie}): 

\begin{enumerate}[leftmargin=*,noitemsep,label=\textbf{Step \arabic*.}]
    \item \textbf{Level-1 screening} — Annotators categorize the paragraph into one of three high-level labels: 
    \emph{W (Affirmative)}, \emph{R (Prohibitive)}, or \emph{N (Neutral)}. 
    \emph{W} represents all affirmative directives, including \emph{B (Authorizing)}, \emph{C (Flexible)}, \emph{G (Ambiguous)}, and \emph{Y (Pressuring)};
    \emph{R} represents directives indicating prohibition; while \emph{N} is used for paragraphs that contain no directives, consisting solely of descriptive or contextual information.
    
    \item \textbf{Level-2 refinement} — If a paragraph is labeled as \emph{W (Affirmative)} at Level-1, 
    Annotators will specify the exact color using one of the following codes: \emph{B (Authorizing)}, \emph{C (Flexible)}, \emph{G (Ambiguous)}, or \emph{Y (Pressuring)}. 
    The paragraph is labeled as \emph{B (Authorizing)} when it demonstrates authorization, 
    \emph{C (Flexible)} when it encourages flexible means to implement, 
    \emph{G (Ambiguous)} when the directive is ambiguous and open to interpretation, 
    and \emph{Y (Pressuring)} when it demands immediate action with a sense of urgency or coercion. 
    If directives cannot be classified into any of the four specified colors, the paragraph will be labeled as \emph{U (Unsure)} to indicate uncertainty.

    \item[\textbf{Label \emph{U}}] We reserve \emph{U (Unsure)} for genuinely unresolved cases after applying the above rules. 
    \emph{U (Unsure)} works as a last resort and will only be used when necessary.
    
\end{enumerate}

This rigorous, two-stage protocol balances theoretical nuance with operational reliability, enabling the creation of a high-quality training set for subsequent NLP models.

\section{Composition of the CAPC-CG}

The CAPC-CG corpus is organized as a relational database designed to facilitate multi-level analysis. It adopts a two-level structure across three document types—Laws, Regulations, and Rules—with each type represented by a pair of tables: documents\_* tables that store document-level metadata and paragraphs\_* tables that store paragraph-level text together with annotation outputs. Within each pair, the two tables are linked through DocumentID, which functions as the primary document identifier and establishes a one-to-many relationship between source documents and their constituent paragraphs. Basic descriptive statistics are reported in Appendix~\ref{app:additional_stats}, while the corpus creation workflow, annotation process, and segmentation pipeline are detailed in Appendix~\ref{app:roadmap}, Appendix~\ref{app:codebook}, and Appendix~\ref{sec:segmentation}. The full field-level schema is provided in Appendix~\ref{app:db_schema}.

\section{Prediction of Directives Color}

While our manual annotation yielded highly accurate data, it is not feasible to scale this method to more than 3.3 million paragraph units. For datasets of such immense size, it is essential to automate annotation and classification \cite{tan2024large}. A few recent studies~\cite{sun2025empowering} attempted to annotate data using LLM APIs with prompt-based instructions; however, existing studies and our own experiments demonstrate that this approach can be highly unreliable \cite{dong2022survey, baumann2025}. In our experience, even three researchers with knowledge of Chinese policymaking required a rigorously designed codebook and extensive training to achieve a satisfactory level of agreement \cite{interarticle}. After comparing multiple strategies, we find that fine-tuning LLMs on a carefully constructed, human-annotated gold-standard dataset provides the most reliable and efficient solution. In this section, we present our experiments on policy classification using LLMs under different settings and report baseline results that both validate the reliability of our dataset and establish a benchmark for future research. This approach offers a cost-effective, scalable, and robust pathway for applying LLMs to similar tasks in computational social science.

\subsection{Experimental Setup}
Our aim was to create an automated pipeline for classifying policy paragraphs according to our two-level annotation schema. All experiments were conducted using our human-annotated gold-standard dataset, which contains 6,000 samples, divided into training (80\%) and validation (20\%) sets. For a final evaluation, we used a separate test set of 1,901 samples, which was collaboratively constructed and validated by our three domain experts.

Our experimental process followed four stages. First, we conducted extensive prompt engineering to find the most effective instructions for our classification tasks. The optimal prompt, detailed in Appendix~\ref{app:prompt}, was identified through iterative testing on OpenAI models and applied in all subsequent experiments. Second, we established baselines using traditional machine learning models (SVM and XGBoost)~\cite{cortes1995svm, chen2016xgboost} and benchmarked a range of state-of-the-art LLMs for the zero-shot Level-1 classification task to identify a model that balances performance and cost-efficiency. We additionally benchmarked representative open-weight encoder-only and decoder-only models under supervised fine-tuning, including BERT-base-Chinese, Qwen2.5-7B, and Llama-3-8B; the full benchmark is reported in Appendix~\ref{app:open_weight_benchmark}. Third, using the selected model, we compared zero-shot, few-shot, and fine-tuning methods to determine the most effective strategy \cite{brown2020languagemodelsfewshotlearners}. Finally, we applied this optimal strategy to the more granular Level-2 (Color) classification task.

Fine-tuning details are reported in Appendix~\ref{app:reproducibility}. In brief, we used the OpenAI fine-tuning API with \texttt{n\_epochs=3}, \texttt{batch\_size=auto}, and \texttt{learning\_rate\_multiplier=auto}; the prompts used for training and inference are provided in Appendix~\ref{app:prompt}.

\subsection{Results and Analysis}

\subsubsection{Level-1 Directive Classification (R/W/N)}
For the main text, we focus on the selected zero-shot baselines most relevant to model selection for scalable inference, while Appendix~\ref{app:open_weight_benchmark} reports the broader benchmark with supervised open-weight models. Within the zero-shot API setting in Table~\ref{tab:zeroshot_comparison}, GPT-4o-mini~\cite{openai2024gpt4omini} attains the best overall metrics (Kappa = 0.618; Acc. = 0.793; M-F1 = 0.692) at substantially lower cost and latency. Balancing throughput, responsiveness, and unit cost, we therefore adopt GPT-4o-mini for subsequent experiments and large-scale inference.

Next, we compared the performance of zero-shot, few-shot (k=3/6), and fine-tuning methods using GPT-4o-mini. The results, presented in Table~\ref{tab:method_comparison}, unequivocally demonstrate the superiority of fine-tuning. Our fine-tuned GPT-4o-mini model achieves a Cohen's Kappa of 0.841 and a Macro-F1 score \cite{SOKOLOVA2009427} of 0.844, substantially outperforming the best few-shot baseline (Kappa = 0.686). This represents a 22.6\% improvement in Kappa, with particularly strong gains on the minority `R' class, validating our strategy of leveraging the gold-standard dataset for model specialization. Appendix~\ref{app:open_weight_benchmark} further shows that fine-tuned BERT-base-Chinese is a strong open-weight Level-1 baseline (Kappa = 0.817; M-F1 = 0.846), but fine-tuned GPT-4o-mini still achieves the highest agreement with expert labels overall.

\subsubsection{Level-2 Color Classification (B/C/G/Y)}
Building on the success of our Level-1 model, we applied the same fine-tuning methodology to the more nuanced four-class Level-2 task, which classifies affirmative directives into specific colors (B, C, G, Y). The fine-tuned GPT-4o-mini model continued to exhibit high performance and reliability on this challenging task. As detailed in Table~\ref{tab:level2_results}, the model achieved an overall accuracy of 87.5\% and a multi-class Cohen's Kappa of 0.833, indicating a strong agreement with our expert annotators. The model performed consistently well across all four categories, demonstrating its capability to capture the subtle distinctions in policy language required for our fine-grained analysis. This result confirms that our fine-tuning approach is robust and effective for both levels of our annotation schema. Appendix~\ref{app:open_weight_benchmark} shows that BERT-base-Chinese is again the strongest open-weight model on this stage, but it still trails fine-tuned GPT-4o-mini (Kappa = 0.800 vs. 0.833; Acc. = 0.850 vs. 0.875), especially on \textit{Grey (Ambiguous)} cases. Error analysis further shows that Level-1 mistakes mainly arise when the model confuses affirmative directives with prohibitions in passages that authorize the state to sanction others. For Level-2, most errors concentrate on the distinction between \textit{Charcoal (Flexible)} and \textit{Grey (Ambiguous)}, reflecting the pragmatic difficulty of separating explicit encouragement of flexibility from strategic hedging. This pattern suggests that the task is not saturated: the large gap between zero-shot and fine-tuned performance shows that expert supervision remains essential, and ambiguity detection in Level-2 continues to leave substantial room for future work. Failure cases are reported in Appendix~\ref{app:benchmark_error}.

\begin{table}[htbp]
\centering
\small 
\setlength{\tabcolsep}{3pt} 
\begin{tabular}{lcccccc}
\toprule
\textbf{Model} & \textbf{Kappa} & \textbf{Acc.} & \textbf{R-F1} & \textbf{W-F1} & \textbf{N-F1} & \textbf{M-F1} \\
\midrule
XGBoost  & 0.153 & 0.487 &0.233 & 0.538 & 0.488 & 0.420 \\
SVM     & 0.175 & 0.508 &0.260 & 0.539 & 0.527 & 0.442 \\
DeepSeek-V3     & 0.481 & 0.733 & 0.438 & 0.613 & 0.815 & 0.622 \\
Qwen3-235B    & 0.570 & 0.780 & 0.561 & 0.705 & 0.834 & 0.700 \\
\textbf{GPT-4o-mini} & \textbf{0.618} & \textbf{0.793} & \textbf{0.482} & \textbf{0.721} & \textbf{0.873} & \textbf{0.692} \\
\bottomrule
\end{tabular}
\caption{Selected zero-shot baselines for the Level-1 task. Full benchmark results, including supervised open-weight models, are reported in Appendix~\ref{app:open_weight_benchmark}.}
\label{tab:zeroshot_comparison}
\end{table}

\begin{table}[htbp]
\centering
\small 
\setlength{\tabcolsep}{2.8pt} 
\begin{tabular}{lcccccc}
\toprule
\textbf{Method} & \textbf{Kappa} & \textbf{Acc.} & \textbf{R-F1} & \textbf{W-F1} & \textbf{N-F1} & \textbf{M-F1} \\
\midrule
Zero-shot        & 0.618 & 0.793 & 0.482 & 0.721 & 0.873 & 0.692 \\
Few-shot (k=3)   & 0.686 & 0.830 & 0.489 & 0.786 & 0.898 & 0.724 \\
Few-shot (k=6)   & 0.659 & 0.816 & 0.512 & 0.763 & 0.883 & 0.719 \\
\textbf{Fine-tuning} & \textbf{0.841} & \textbf{0.910} & \textbf{0.677} & \textbf{0.905} & \textbf{0.952} & \textbf{0.844} \\
\bottomrule
\end{tabular}
\caption{Comparison of methods for Level-1 classification using the \textbf{GPT-4o-mini} model. Fine-tuning shows substantial improvement.}
\label{tab:method_comparison}
\end{table}

\begin{table}[htbp]
\centering
\begin{tabular}{p{5.9cm} p{0.8cm}}
\toprule
\textbf{Metric} & \textbf{Score} \\
\midrule
Cohen's Kappa (4-class) & 0.833 \\
Overall Accuracy & 87.5\% \\
\midrule
\multicolumn{2}{l}{\textit{Accuracy per Class:}} \\
\quad Class `B' (Authorizing) & 92.0\% \\
\quad Class `C' (Flexible) & 85.0\% \\
\quad Class `G' (Ambiguous) & 84.0\% \\
\quad Class `Y' (Pressuring) & 89.0\% \\
\bottomrule
\end{tabular}
\caption{Selected Level-2 results for the fine-tuned GPT-4o-mini model. Full open-weight benchmark results are reported in Appendix~\ref{app:open_weight_benchmark}.}
\label{tab:level2_results}
\end{table}

\section{Data Analysis}
Our corpus enables nuanced analyses of Chinese policy communication. To demonstrate its utility, we present two preliminary findings.

\begin{figure*}[htbp!]  
  \centering
  \includegraphics[width=\textwidth,
                   height=.50\textheight,
                   keepaspectratio]{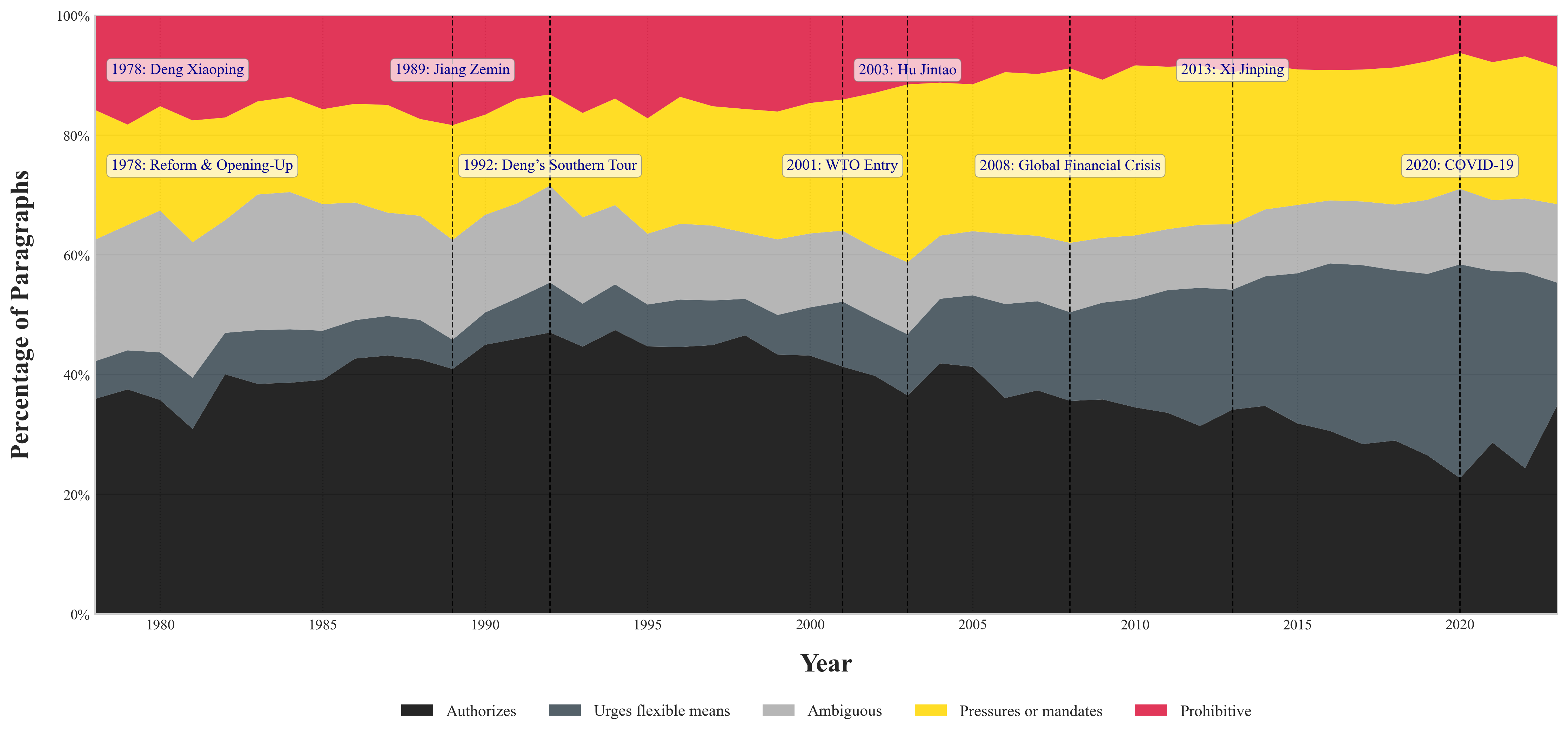}
  \caption{Temporal Trend of Policy Signals after the Reform \& Opening-Up (1978--2023)}
  \label{fig:color_trend}
\end{figure*}

 We begin with the temporal evolution. Figure~\ref{fig:color_trend} shows that certain patterns have remained quite consistent: for example, \textit{Red (Prohibiting)} directives make up only a small portion, compared to \textit{Black (Authorizing)}—especially after the Reform and Opening-Up period began in 1978. Moreover, this pattern appears to be influenced more by major events, such as economic shifts or public health crises, than by leadership transitions.

\begin{figure}[htbp!]
    \centering
    \includegraphics[width=1\linewidth]{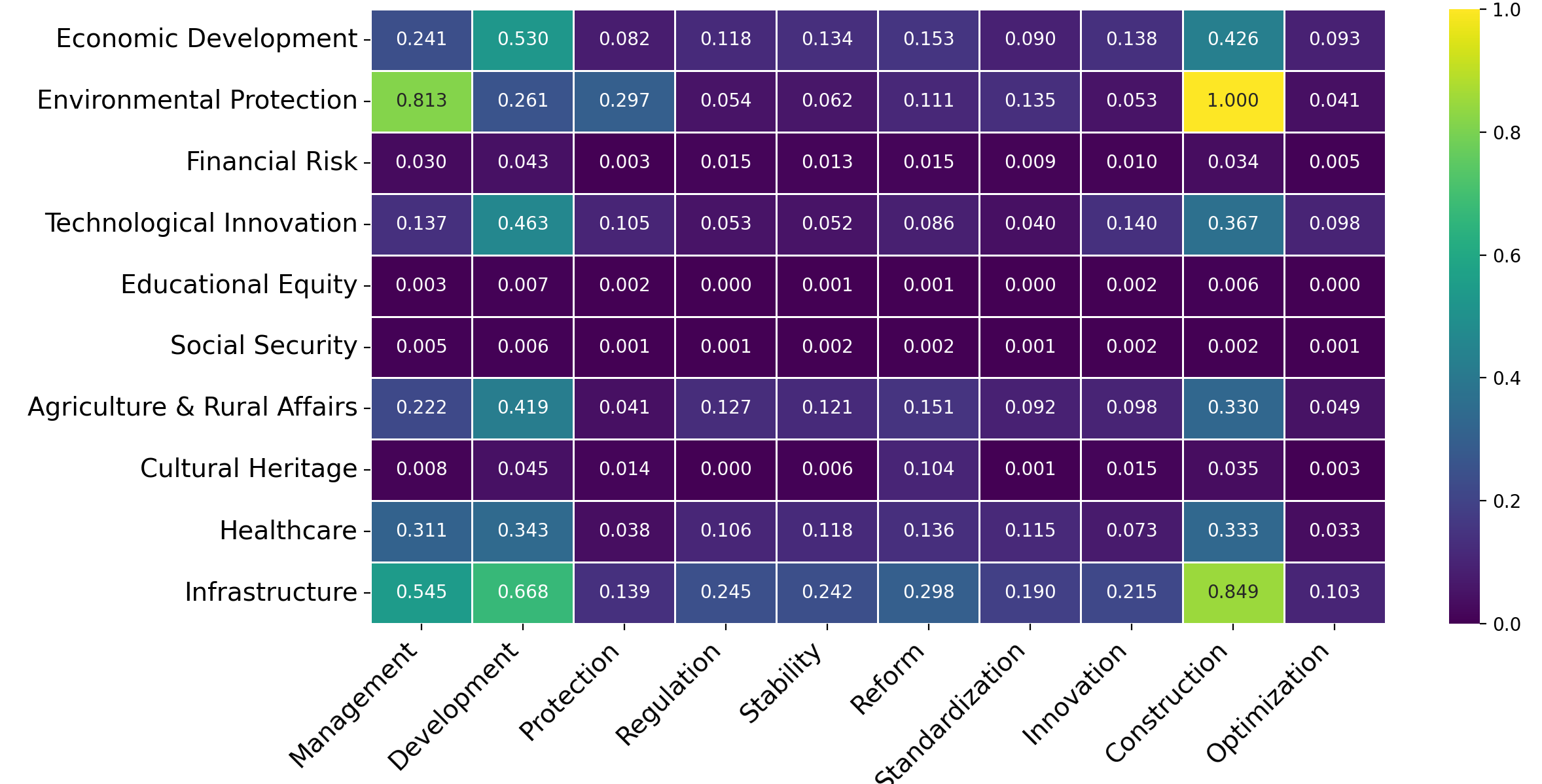}
    \caption{The distribution of action-oriented topics across different policy domains}
    \label{fig:heatmap}
\end{figure}

 Figure~\ref{fig:heatmap} visualizes the topic distribution derived from our LDA model through heatmap, revealing a strong thematic concentration for certain policy domains (y-axis) across several action-oriented topics (x-axis). Notably, the `Environmental Protection' corpus is almost entirely dominated by the `Construction' topic, while the `Infrastructure' domain is also heavily characterized by `Construction', as well as `Development' and `Management.' This concentration suggests a primary textual focus on tangible, project-based activities within these policy areas. In stark contrast, social policy domains such as `Educational Equity' and `Social Security' exhibit negligible weights across all these topics, indicating that their discourse is thematically distinct and likely represented by other latent topics in the model.

\begin{figure}[htbp!]
    \centering
    \includegraphics[width=1\linewidth]{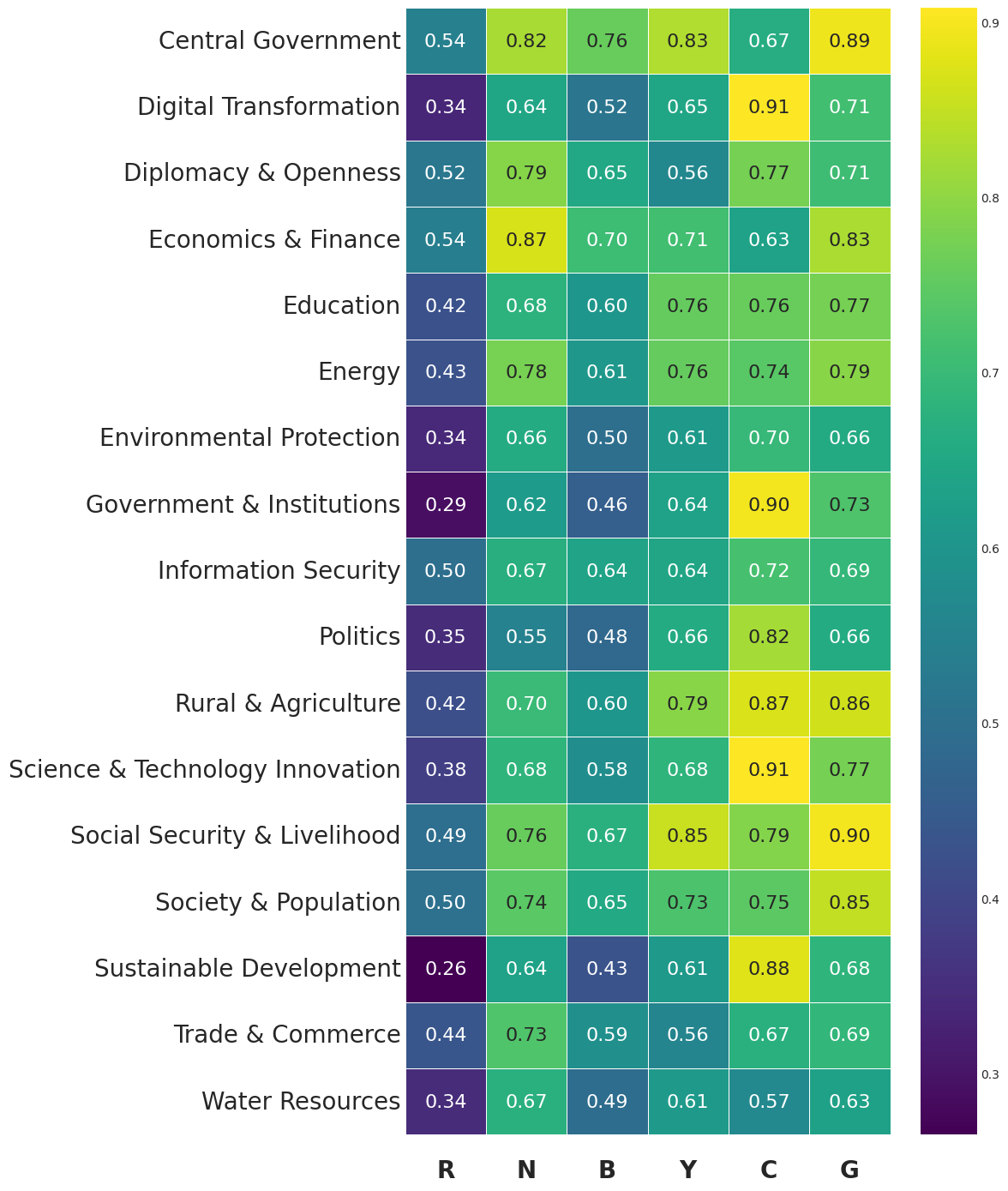}
    \caption{Heatmap of governmental intents across policy domains}
    \label{fig:heatmap_color}
\end{figure}

Figure~\ref{fig:heatmap_color} visualizes the distribution of governmental intents (x-axis), as defined in Table~\ref{tab:signals}, across 17 policy domains. The analysis reveals that specific intents are strongly correlated with particular types of domains. Notably, the intent to ``urge flexible means'' (C) is almost exclusively dominant in technology-driven domains such as `Digital Transformation' (0.91) and `Science \& Technology Innovation' (0.91), suggesting a governance model that prioritizes local experimentation.

A useful counterexample is Water Resources, where the intent to ``urge flexible means'' (C) is comparatively weak. Unlike technology- or innovation-oriented domains, water governance often depends on clearly specified targets, implementation procedures, technical standards, and administrative requirements. This suggests that the policy style in this domain relies less on experimentation or differentiated local pathways, and more on structured implementation through defined goals and concrete execution methods.

\section{Future Directions}

As a new and foundational resource, our corpus can open up at least three avenues for future research. 

\subsection{CAPC-CG as Foundation for Multi-Level Policy Communication}
\label{data expansion}
As an essential first step, the \textbf{CAPC-CG} corpus covers central-level documents. Future work can collect directives issued by provincial and municipal governments to enable analysis of multi-level policy communication and diffusion. In future, we also hope to integrate supplementary materials such as leaders' speeches and state media commentary. 

\subsection{Expert-Directed LLM Annotation Method}
\label{china-policy-nlp}
Our experiments highlight the limitations of state-of-the-art LLMs in capturing the nuances of Chinese policy directives, particularly variations in clarity and ambiguity. To address this, we introduce and validate a new method: expert-directed LLM annotation. Rather than relying on prompt-based annotation alone, our approach integrates expert-designed codebooks, rigorous annotator training, and fine-tuning on a gold-standard labeled set to capture varied and ambiguous policy signals. Unlike prior work that treats LLMs as substitutes for expert human annotators, our method treats expert annotation as the foundation and uses LLMs to scale this system across large corpora. 

\subsection{Adaptive Political Economy: From Paradigm to Operationalization}
\label{computational-social-science}
Our dataset serves as a high-resolution empirical resource for adaptive political economy (APE) \citep{ang2024ape}, a paradigm for studying political economies as complex adaptive systems. APE develops concepts, theories, and methods that illuminate complex social features such as adaptation and uncertainty, rather than assuming them away. Applied to organizational analysis, APE contrasts conventional principal-agent models that assume central authorities have clear goals and seek to control agents \citep{kiewiet1991logic}. Adaptive policy communication theorizes communication in a context where leaders calibrate discretion by adjusting their mixture of signals and without always having fixed policy preferences. 

APE can include but is not limited to qualitative analysis; case studies serve to illustrate a theory before it is operationalized at scale \citep{ang2016china}. Our study demonstrates that APE, as a paradigm, can generate new lines of inquiry (how leaders communicate under complexity), theory (adaptive policy communication), typology (five policy signals), measurement (theory-driven, human-LLM annotation), and novel data (CAPC-CG), which can inspire more data collection. 




\section{Conclusion}

This paper introduces and validates \textbf{CAPC-CG}, which, to our knowledge, is the most comprehensive paragraph-level corpus of Chinese policy directives for NLP research, built around a five-color typology of policy signals. By innovating a theory-driven annotation method and leveraging fine-tuned LLMs, we constructed a novel, computable dataset. 

We hope this dataset will enable studies of policy language across political systems and the development of more advanced models. We show how political science theory can be operationalized at scale using cutting-edge methods in NLP, positioning CAPC-CG as a foundation for research in adaptive and comparative policy communication.


\section*{Limitations}

We acknowledge the following limitations, which also point to avenues for future research.

First, our corpus is subject to the inherent challenges of archival availability and opacity in contemporary China. Although we have systematically compiled the most comprehensive collection possible from public and official databases, the complete historical record is not fully accessible. Notably, numerous documents from politically sensitive periods, such as the Cultural Revolution (1966-1976), are intentionally withheld from the public domain or remain in undigitized archives. Consequently, while our dataset represents the fullest publicly available record, it cannot capture the entirety of the CCP's internal policy discourse during certain historical junctures. Our analysis should therefore be understood as pertaining to China's public-facing policy landscape.

Second, our gold-standard dataset was annotated by a small, dedicated team of three domain experts rather than a large crowdsourced workforce. This decision was necessitated by the significant domain knowledge required to interpret Chinese policy language and the complexity of our annotation schema, which demanded an intensive, multi-month training period to ensure reliability. We deliberately prioritized the quality, depth, and consistency of annotations over the sheer quantity of annotators. The high inter-annotator agreement achieved (Fleiss' Kappa = 0.864) validates this expert-driven approach and the reliability of our data, though we acknowledge that a larger pool of annotators could have captured a wider range of interpretations.

Finally, although Appendix~\ref{app:open_weight_benchmark} extends the evaluation to several representative open-weight encoder-only and decoder-only models, our experiments are still not exhaustive. Our primary goal was to construct and publicly release the \textbf{CAPC-CG} corpus, and the modeling experiments were designed mainly to illustrate its utility rather than to perform a fully comprehensive benchmarking exercise. Evaluating a wider array of open-weight models, adaptation strategies, and hyperparameter settings will further strengthen the baseline. We hope that our released dataset will serve as a standard testbed for such comparative analyses, supporting more extensive exploration by the computational social science and NLP community.

\section*{Acknowledgments}
This material is based upon work supported by the National Science Foundation under Award No. 2316967.


\bibliography{custom}

\clearpage
\appendix

\section{Appendix Roadmap and Corpus Workflow}
\label{app:roadmap}
\begin{figure*}[t]
    \centering
    \includegraphics[width=0.95\textwidth]{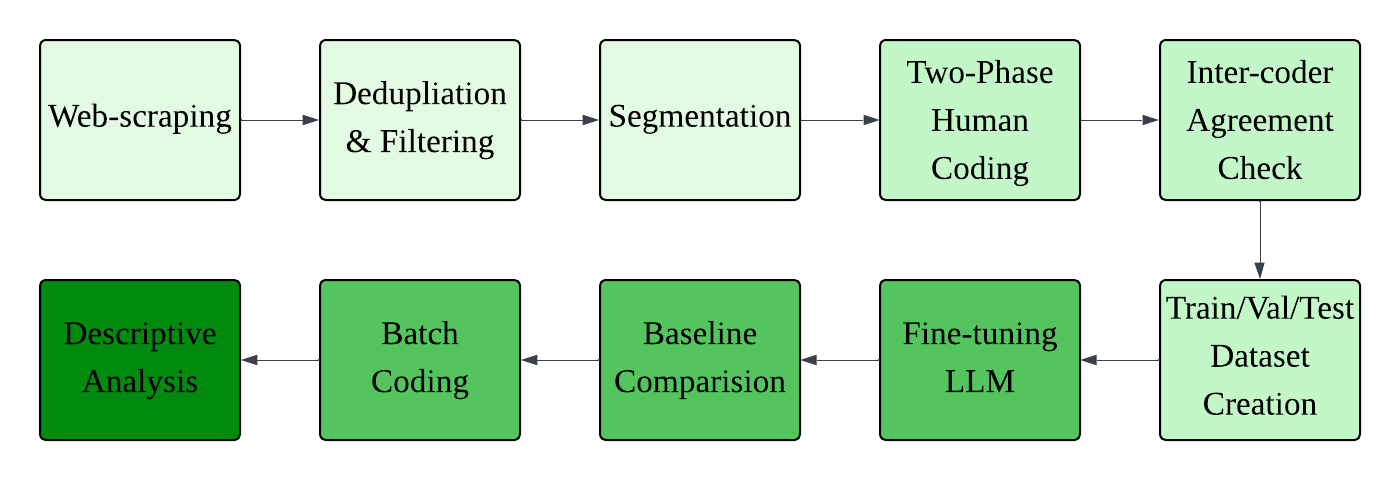}
    \caption{Overall Workflow}
    \label{fig:Workflow}
\end{figure*}

This section provides a roadmap to the supplementary materials and a high-level overview of the end-to-end CAPC-CG construction process. Appendix~\ref{app:codebook} summarizes the annotation codebook, Appendix~\ref{sec:segmentation} details the segmentation pipeline, Appendix~\ref{app:prompt} reports the prompts used for annotation and model training, and Appendix~\ref{app:reproducibility} records the core fine-tuning settings. The later sections provide extended benchmark results, bilingual examples, benchmark error analysis, schema documentation, topic-modeling details, and additional descriptive statistics. Figure~\ref{fig:Workflow} presents the overall workflow that connects these components.

\section{Codebook and Annotation Process}
\label{app:codebook}

\subsection{Purpose and Scope}
To enhance comprehension, this appendix concisely describes our codebook and workflow for annotating \emph{policy directives} in the form of three types of central policy documents, issued between 1949 and 2023: (i) National Laws, (ii) Administrative Regulations, and (iii) Ministerial Rules.

\subsection{Theory}
Adaptive policy communication is theory of governance under complexity, proposed by \citet {ang2016china}, which explains how leaders steer policy implementation by combining clear and ambiguous instructions. Building on \citet{ang2023ambiguity}, we annotate policy directives around a five-color typology:
\begin{itemize}
  \item \textbf{Red (R)}: Clear prohibition; often accompanied by penalties for violations.
  \item \textbf{Black (B)}: Clear authorization, approval, or assignment of responsibility. 
  \item \textbf{Yellow (Y)}: Clear mandates with urgency or intense pressure; often accompanied by binding targets, deadlines, or performance reviews.
  \item \textbf{Charcoal (C)}: Clear mandates to implement policies flexibly (e.g. experiments and local adaptation).
  \item \textbf{Grey (G)}: Ambiguous language or hedging policy intentions
\end{itemize}

In addition to this five-color typology, we introduce the label \textbf{N} to designate paragraphs that contain no actionable directives. These paragraphs typically include background information, ideological slogans, personnel notices, logistical notes, instructions targeting non-state actors.

\subsection{Coding Principles}
\begin{enumerate}[leftmargin=*]
  \item \textbf{Recipient principle}: All directives should be read from the perspective of state actors. Instructions directed solely at individuals, enterprises, or social groups without assigning bureaucratic tasks are not considered directives.
  \item \textbf{Clarity vs. specificity}: Code for clarity of intent, not specificity of detail. A directive may express a clear purpose while remaining vague in its implementation, or contain extensive details yet lack a clear underlying intent.
  \item \textbf{Avoid keyword-only tagging}: Keywords like \textit{must}, \textit{flexible}, or \textit{forbid} can serve as helpful cues, but they should never be used mechanically or treated as the sole criteria.
\end{enumerate}


\subsection{Step-by-Step Workflow}
This section outlines our workflow in a two-round annotation process. 

\paragraph{Step 1: Assign Level 1 Label (W / R / N)}
Determine whether the paragraph belongs to:
\begin{itemize}
    \item W (White): Directive for actions (B, Y, C, G)
    \item R (Red): Prohibition
    \item N (None): No directive
\end{itemize}

\paragraph{Step 2: If W, identify B / Y / C / G / U}

If the paragraph clearly fits one category, label it as R, B, Y, C or G. If the paragraph contains mixed signals, apply the \textbf{Proportion Method}:
\begin{itemize}
    \item Break the paragraph into individual sentences
    \item Assign a color signal to each sentence
    \item Label the entire paragraph based the color signal with the largest proportion
\end{itemize}

\paragraph{Step 3: Use U (Unsure) only when necessary}
\begin{itemize}
    \item Assign U only if the paragraph cannot be confidently labeled after applying the first two steps.
    \item Use U as a last resort. Do not assign U to avoid difficult judgments.
\end{itemize}



\subsection{Data Schema}
For each paragraph, record:
\begin{itemize}
  \item Document ID and Paragraph ID
  \item Level 1 label: W/R/N
  \item Level 2 label: B/Y/C/G/U
  \item Optional notes (brief rationale/phrase)
  \item Recommended metadata: document type, issuing unit, receiving unit (if any), abbreviation/number, title, year.
\end{itemize}

\subsection{Intended Use}
This codebook undergirds reliable human annotation, supports transparency for external researchers and ML collaborators, and facilitates aggregation for downstream modeling.

\section{Segmentation Method}
\label{sec:segmentation}
This appendix provides a comprehensive technical description of the two-stage computational pipeline developed to segment the corpus of Chinese policy documents. The primary objective of this methodology is to partition each document into discrete, semantically coherent, and atomic units of policy prescription. This process is foundational to our research, as it ensures that subsequent quantitative analyses are based on consistent and replicable data, free from the non-deterministic artifacts of purely generative approaches.

The core challenge this pipeline solves is the inherent unreliability of using an LLM alone for segmentation. A standalone LLM exhibits non-deterministic behavior on hierarchical text, making it unsuitable for research requiring replicable data processing. Our hybrid approach overcomes this by using the LLM solely for structural identification, while a deterministic script performs the final segmentation.

\subsection{Corpus Description}
The corpus to which this pipeline was applied consists of 337,038 Chinese policy documents issued between 1949 and 2023. The documents fall into three main categories:
\begin{itemize}[leftmargin=*]
\item Administrative Regulations: 11,621 documents
\item Ministerial Rules: 322,280 documents
\item National Laws: 3,137 documents
\end{itemize}

\subsection{Cost Optimization}
A central design goal was to mitigate API costs. This was achieved by architecting the pipeline to minimize the expensive data transfer in the LLM's output. To fully appreciate the innovation, we contrast our implemented pipeline with a standard, un-optimized approach.

\subsubsection{Hypothetical un-optimized pipeline}
A standard approach would involve sending the raw document text to the LLM and instructing it to return the full text with inserted segmentation tags. The process is as follows:
\begin{enumerate}[leftmargin=*]
\item\textbf{API Call}: Send the entire raw text of document.txt to the LLM.
\item\textbf{LLM Task}: The LLM internally processes the text and embeds XML tags (e.g., <L1>...) directly into the content.
\item\textbf{LLM Output}: The LLM returns the complete, modified text as its response.
\end{enumerate}
\paragraph{Cost analysis} At the project's pricing of \$0.20/1M input and \$0.80/1M output tokens, this model is prohibitively expensive. The output token count is nearly identical to the input count, and output tokens are four times more costly. The total cost for a single document would be approximately: Cost un-optimized \(\approx\) (Tokens input \(\times\) \$0.20) + (Tokens output \(\times\) \$0.80)

Since Tokens output \(\approx\) Tokens input, the cost is dominated by the output, making the process five times more expensive than just reading the input.

\subsubsection{Implemented optimized pipeline}
Our implemented pipeline fundamentally changes the LLM's task from "text editor" to "structural analyst," shifting the workload locally to reduce API costs. The process is as follows:
\begin{enumerate}[leftmargin=*]
\item\textbf{Local Pre-processing}: A Python script first adds <line \#> markers to the document text. This step happens locally and incurs no API cost.
\item\textbf{API Call}: Send the numbered text to the GPT-4.1-mini-2025-04-14 model via the OpenAI Batch API.
\item\textbf{LLM Task}: The LLM's only task is to identify structural elements and output a compact JSON file listing labels and their corresponding line numbers.
\item\textbf{Local Reconstruction}: A local script merges the LLM's JSON output with the original text file to create the XML-tagged document.
\end{enumerate}
\paragraph{Cost analysis} This architecture transforms the cost equation. The output is now a very small JSON object, typically only 5-10\% of the input token size. The total cost is: Cost optimized \(\approx\) (Tokens input \(\times\) \$0.20) + (Tokens JSON \(\times\) \$0.80)

This workflow nearly eliminates the expensive output token cost, making the entire process vastly more economical.

\subsubsection{Results}
The cost-optimization strategy was exceptionally effective. The total expenditure for processing the corpus was \$294.99. In contrast, the un-optimized pipeline, with its 1:1 input-to-output token ratio, would have incurred an estimated cost of \$1,061.30. Our implemented pipeline therefore achieved a verifiable cost reduction of 72.20\%.
The pipeline was validated through an iterative process of expert review. 

\subsection{Pipeline Workflow}
\begin{figure}[htbp!]
    \centering
    \includegraphics[width=1\linewidth]{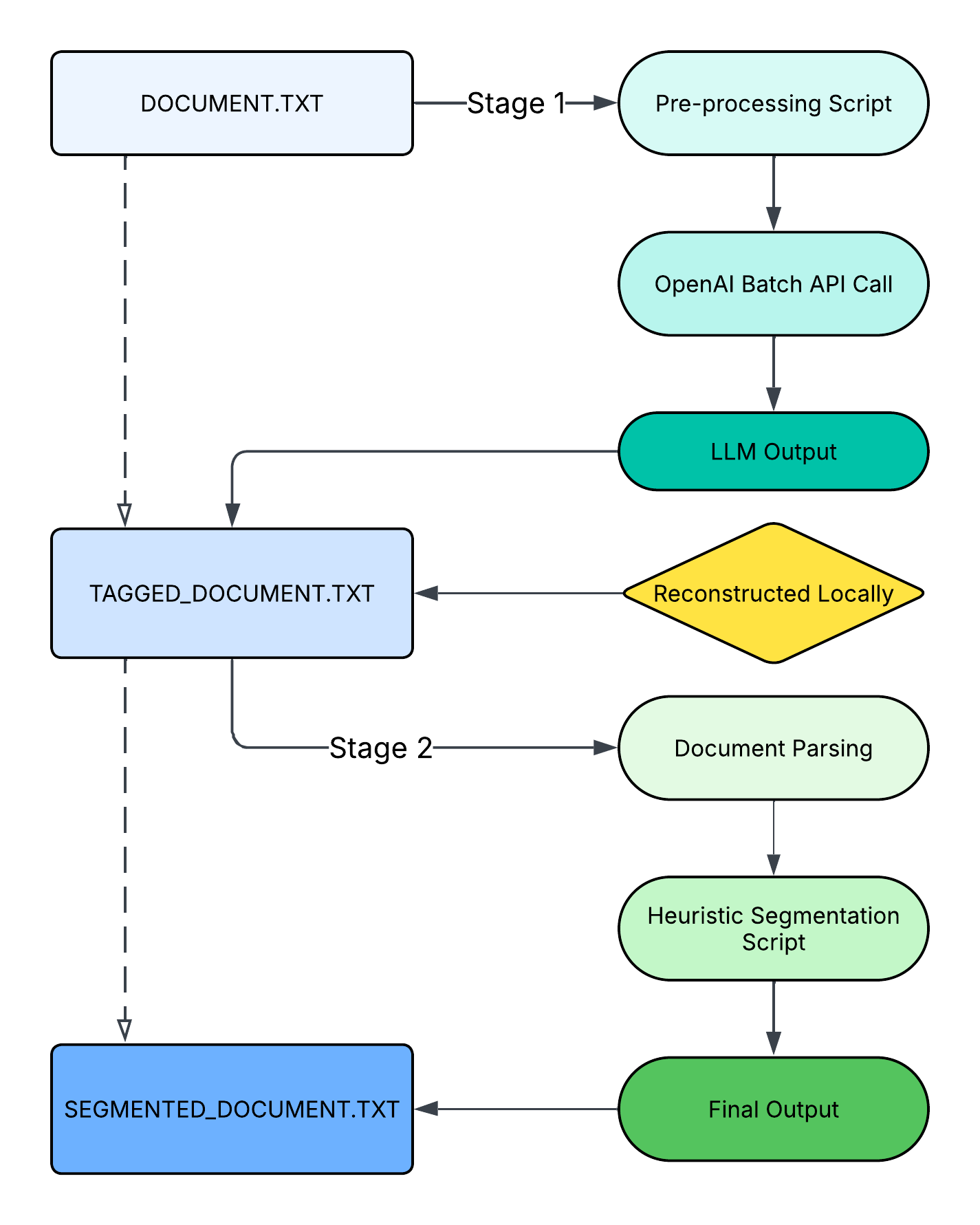}
    \caption{Segmentation Workflow}
    \label{fig:segment_workflow}
\end{figure}

Figure \ref{fig:segment_workflow} illustrates the end-to-end process, from raw text file to final segmented output.

\subsubsection{Stage 1: LLM-Based Structural Tagging}
This stage is orchestrated by a master Python script leveraging the OpenAI and tiktoken libraries. It prepares files, submits them to the Batch API, and monitors their progress. The core instruction is contained within the system prompt provided to the GPT-4.1-mini-2025-04-14 model:\\

\zh{您是专门用于对中文法律文件进行结构化分析的AI助理。您的任务是识别文本中的特定段落，并根据其内容和结构进行分类。输入文本的每一行都以<}line\#\zh{>作为前缀。
核心目标：隔离层级列表项} (L\#\zh{标签}) \zh{和无意义的样板文本} (NEUTRAL\_CONTENT \zh{标签})\zh{，同时保留有意义的散文段落不进行标记，以供后续分析。}

\zh{输出要求：您的输出必须是一个}JSON\zh{对象。此对象包含一个名为}``segments''\zh{的键，其值为一个数组。数组中的每个对象都应包含：}``label'': (string) \zh{必须是}``L1'', ``L2'', ``L3'', ..., \zh{或} ``NEUTRAL\_CONTENT''\zh{。}``line\_numbers'': (array of integers) \zh{与该标签对应的原始行号列表。}

\zh{标记规则：
层级标签} (L\#): \zh{使用} L1\zh{、}L2 \zh{等标签来标记结构化的、有编号或有字母的列表项。以\ 第\#条\ 开头的列表项是最高层级，必须标记为}L1\zh{。}

\zh{中性内容} (NEUTRAL\_CONTENT): \zh{仅用于标记客观的、程序性的、或不具有任何分析意义或情感色彩的样板文本。包括但不限于：文件标题、文首、目录、人名、引用编号、称呼、结尾落款和日期、简单的项目列举。

不标记的内容: 提供解释、推理、条件或背景信息且不属于列表结构的散文段落，被视为有意义的内容。这些段落必须保持不标记。
最终指令：严格按照上述规则分析提供的文本，并仅返回一个}JSON\zh{对象。不要包含任何额外的解释或评论。}

\subsubsection{Stage 2: Rule-Based Analysis and Segmentation}
The second stage uses a local Python script that relies on the BeautifulSoup library with the lxml-xml parser. This stage performs two distinct, sequential tasks: first, it analyzes the document's content to generate metadata, and second, it applies a deterministic ruleset to perform the final segmentation.
\begin{enumerate}[leftmargin=*]
\item\textbf{Metadata Generation: Heuristic Content Analysis}

Before segmentation, the script first analyzes the tagged document to determine its most substantively important structural layer. This is achieved by scoring the content within each hierarchical layer (e.g., all L1 tags, all L2 tags) based on a set of heuristic parameters. The goal of this step is to produce a single piece of metadata—the "optimal segmentation layer"—which is saved in the header of the output files for potential use in later analysis.

This scoring process is governed by tunable parameters in Table \ref{tab:parameter_weights}.

\item\textbf{Final Segmentation: Deterministic Rule Application}

After the metadata has been generated, a separate function performs the final, deterministic segmentation. This process does not use the heuristic scoring parameters above. Instead, it inserts segment breaks based on the following purely structural rules:
\paragraph{Structural Change} A break is inserted when transitioning between a content tag (e.g., <L1>) and a <NEUTRAL\_CONTENT> tag. This isolates substantive policy from boilerplate.
\paragraph{Hierarchical Ascent} A break is inserted if the hierarchical level of the current tag is the same as or higher than the previous tag (e.g., <L2> to <L2>, or <L3> to <L2>). This rule correctly isolates self-contained articles or major sections.
\paragraph{Punctuation Exception} A break is suppressed if the preceding text block ends with a joining punctuation mark (`\zh{,}', `\zh{；}', `\zh{,}', `\zh{;}') and the next tag is at the same level. This sophisticated rule prevents the fragmentation of compound sentences and related clauses.
\end{enumerate}

\begin{table*}[t]
  \centering
  \renewcommand{\arraystretch}{1.5}
  \small
  \begin{tabular}{p{3.8cm}p{1cm}p{10.8cm}}
        \toprule
        \textbf{Parameter} & \textbf{Value} & \textbf{Rationale} \\
        \hline
        \texttt{COLOR\_DIVERSITY\_WEIGHT} & 3.0 & Emphasizes the presence of action-oriented keywords. \\
        \texttt{LENGTH\_WEIGHT} & 0.5 & Favors longer, more substantive text blocks. \\
        \texttt{LAYER\_PENALTY\_FACTOR} & 1.5 & Penalizes deeper layers (L3, L4), creating a bias for higher-level structure. \\
        \texttt{MIN\_LENGTH\_THRESHOLD} & 15.0 & Minimum character count for a block to be considered score-worthy. \\
        \texttt{CHINESE\_ACTION\_WORDS} & 12.0 & Set of keywords indicating policy instructions or legal mandates: \zh{"必须", "应", "须", "可以", "可", "建议", "宜", "不得", "禁止", "严禁", "立即", "紧急"} \\
        \hline
    \end{tabular}
    \caption{Parameters and Weights Description}
    \label{tab:parameter_weights}
\end{table*}

\subsection{Validation and Performance}
The segmentation pipeline was validated through a detailed manual review process. We randomly selected a set of documents and compared the results of machine-based segmentation and manual segmentation by two Research Assistants with advanced background knowledge in Chinese politics. 

The primary criterion for evaluation was semantic coherence, specifically assessing whether each segment served as a complete and intelligible unit of policy information. Each paragraph was marked as "Yes" for accurate or "No" for inaccurate, allowing for the calculation of accuracy rates. 

The results demonstrated impressive accuracy, with 96.7\% for Ministerial Rules and 98.25\% for National Laws, providing strong evidence of the validation and reliability of our segmentation methodology.

\section{Prompts for Annotation}
\label{app:prompt}

\setlength{\parindent}{0pt} 
\setlength{\parskip}{0pt} 
\subsection{Level-1 (W/R/N) Prompt}
\zh{角色：你是一名``政策信号一级标注员''。}\\

\zh{任务：对给定中文政策文档段落分类标注：}\\

W = \zh{段落中存在对政府/公共部门的肯定性行动指令（包含任何``要/应/负责/建立/加强/推进/组织/指导/监督/落实/制定/完善/开展/实施/安排''等对政府主体布置的任务或授权）。}\\

R = \zh{段落中存在对政府/公共部门的禁止性指令（如``严禁/不得/禁止/停止……（由政府或其部门）''）或对政府及公职人员的违法行为进行处罚。}\\

N = \zh{无指令（仅背景介绍、原则口号、定义释义、程序性条款、日期/效力条款、对非国家主体的要求而未同时对政府布置任务或责任的内容）。}\\

\zh{受众原则（极其重要）：只统计指向政府及其附属公共机构（包括国务院各部委、地方政府、监管机构、事业单位等）的行动/禁止为指令。}\\

\zh{仅限制或要求企业、个人或社会组织，而没有同时给政府安排任务或责任，则记为} N\\

\zh{若同时出现：例如``企业不得……；主管部门要加强监管……''，则本段视为存在对政府的指令，进入} W \zh{或} R \zh{的判断。}\\

\zh{政府/公共部门的定义为中央或地方政府、公职人员或承担了国家分配的任务的附属机构}\\

\zh{不要只看关键词：是否出现``必须/可以/鼓励/不得''等词不等于对应标签，必须结合主语是否为政府/公共机构与句子功能来判断。}\\

\zh{例如：``企业必须符合××条件方可办理''，若仅针对企业且未布置政府任务，则记为} N\\

\zh{输出要求：只输出最终的结果}W/N/R\\

\zh{待判断政策内容如下：}\\

<<\zh{待判断政策段落}>>\\

Role: You are a ``first-level policy signal annotator.''\\

Task: Classify the given Chinese policy document paragraph into one of the following categories:\\

W = The paragraph contains affirmative action directives directed at the government/public sector (including any assignment of tasks, responsibilities, or authorization to government actors, such as ``should / shall / responsible for / establish / strengthen / promote / organize / guide / supervise / implement / formulate / improve / carry out / execute / arrange,'' etc.).\\

R = The paragraph contains prohibitive directives directed at the government/public sector (e.g., ``strictly prohibit / must not / forbidden / stop ...'' by government or its departments), or imposes penalties on illegal actions by government officials or public institutions. \\

N = No directive (e.g., background information, general principles or slogans, definitions, procedural clauses, date/effectiveness clauses, or requirements directed only at non-state actors without assigning tasks or responsibilities to the government).\\

Audience Principle (critically important): Only count actions or prohibitions directed at the government and its affiliated public institutions (including central ministries, local governments, regulatory agencies, and public service units) as directives.\\

If the paragraph only restricts or requires enterprises, individuals, or social organizations, without assigning any task or responsibility to the government, label it as N.\\

If both appear, e.g., ``Enterprises must not ...; supervisory authorities should strengthen oversight ...'', then the paragraph is considered to contain a government directive and should be classified as W or R.\\

Definition of Government/Public Sector: Central or local governments, public officials, or affiliated institutions that undertake state-assigned responsibilities.\\

Important Instruction: Do not rely solely on keywords. The presence of words like ``must / may / encourage / must not'' does not determine the label. You must consider whether the subject is a government/public institution and the function of the sentence.\\

Example: ``Enterprises must meet certain conditions before approval''; if it only targets enterprises and does not assign any task to the government, label it as N.\\

Output Requirement: Output only one label: W / N / R\\

Policy paragraph to classify:\\

<Policy paragraph>\\

\subsection{Level-2 (B/Y/C/G) Prompt}
\zh{角色：你是一名``政策信号一级标注员''。}\\

\zh{任务：对给定的中文政策文档段落进行分类标注, 该段落已经被判定为含有肯定性政策指令，请你进一步判断它属于以下哪一类信号：}\\

B\zh{（}Black\zh{，你可以）：明确同意、批准或中性部署（如``建立/推进/构建/实现''），没有强制性、没有因地制宜的灵活性、也不含糊。}\\

Y\zh{（}Yellow\zh{，你必须）：带有明显压力或紧迫性，要求全面落实或严格执行，常出现``抓紧/尽快/及时/必须/严格执行/全面落实''等；纳入考核、问责、审计也属于} Y\zh{。注意：如果``必须''只是作为条件或资格要求（例如``必须符合条件方可申请''），则不算} Y\zh{。}\\

C\zh{（}Charcoal\zh{，你应当灵活地做）：指令明确，但强调 ``因地制宜、多种形式、试点示范、先行先试、借鉴与反馈'' 等灵活路径或差异化方式。}\\

G\zh{（}Grey\zh{，你可能做也可能不做）：含糊、模糊或保留条件的指令，常见表达包括``在…前提下'' ``适度'' ``具备条件的'' ``可'' ``鼓励探索'' ``逐步推进'' ``协商解决''等，主体或强度不清晰。}\\

\zh{输出要求：只输出最终的结果}B/Y/C/G\\

\zh{待判断政策内容如下：}\\

<<\zh{待判断政策段落}>>\\

Role: You are a ``first-level policy signal annotator.''\\

Task: Classify the given Chinese policy document paragraph into one of the following categories.
The paragraph has already been identified as containing an affirmative policy directive. Your task is to further determine which type of signal it represents:\\

B (Black, ``you can'') = Explicit approval, endorsement, or neutral deployment (e.g., ``establish / promote / build / implement''), without strong enforcement pressure, no flexibility, and no ambiguity.\\

Y (Yellow, ``you must'') = Directive with clear pressure or urgency, requiring strict or comprehensive implementation. Common expressions include ``grasp firmly / as soon as possible / promptly / must / strictly implement / fully implement.'' Inclusion in evaluation, accountability, or auditing also counts as Y.
Important: If ``must'' is used only as a condition or eligibility requirement (e.g., ``must meet conditions to apply''), it does not count as Y.\\

C (Charcoal, ``you should do flexibly'') = Directive is clear, but emphasizes flexible implementation paths, such as ``adapt to local conditions / multiple forms / pilot programs / experimentation / learn and adjust / feedback mechanisms.''\\

G (Grey, ``you may or may not do'') = Vague, ambiguous, or conditional directives. Common expressions include ``under the premise that...'', ``moderately'', ``where conditions permit'', ``may'', ``encourage exploration'', ``gradually promote'', and ``resolve through consultation.'' In such cases, the subject or level of enforcement is unclear. \\

Output Requirement: Output only one label: B / Y / C / G\\

Policy paragraph to classify:\\

<Policy paragraph>


\section{Reproducibility Details}
\label{app:reproducibility}

\paragraph{GPT-4o-mini fine-tuning}
We fine-tuned GPT-4o-mini using the OpenAI fine-tuning API with \texttt{n\_epochs=3}, \texttt{batch\_size=auto}, and \texttt{learning\_rate\_multiplier=auto}. The gold-standard annotation set contains 6,000 examples and was split 80/20 into training and validation subsets. Final evaluation was conducted on a held-out test set of 1,901 examples.

For both training and inference, we used the prompt templates reported in Appendix~\ref{app:prompt}. The model was required to return exactly one valid label per instance. We normalized minor output-formatting variants in post-processing so that predictions mapped cleanly onto the valid label sets \{W, R, N\} for Level-1 and \{B, C, G, Y\} for Level-2.

\paragraph{Extended open-weight benchmark configuration}
The broader benchmark reported in Appendix~\ref{app:open_weight_benchmark} used a common local training setup across supervised baselines. Unless otherwise noted, models were trained with AdamW, maximum sequence length 512, \texttt{warmup\_ratio=0.1}, \texttt{weight\_decay=0.01}, and early stopping with patience 2 on macro-F1. Traditional ML baselines used TF-IDF features with 1--2 grams and a 5,000-feature cap; the SVM baseline used a linear kernel with \texttt{C=1}, and XGBoost used 200 trees with maximum depth 6.

\paragraph{Open-weight model settings}
For BERT-base-Chinese, we used full-parameter fine-tuning with learning rate \texttt{2e-5}, batch size 16, and 4 epochs. For Qwen2.5-7B and Llama-3-8B, we used QLoRA with 4-bit NF4 quantization, LoRA rank \texttt{r=16}, \texttt{alpha=32}, \texttt{dropout=0.05}, target modules \texttt{[q,k,v,o]\_proj}, learning rate \texttt{1e-4}, effective batch size 16, and 3 epochs. For the Level-2 local benchmark, we used bf16 training and gradient checkpointing for the QLoRA models. Local Level-2 fine-tuning used the paragraph text as model input and the gold label as target, omitting the system prompt contained in the original JSONL files.

\paragraph{Benchmark-specific data splits}
The extended benchmark relies on benchmark-specific splits in addition to the main GPT-4o-mini training setup above. For Level-1, the open-weight benchmark used 600 balanced training samples, 300 balanced validation samples, and a held-out 1,901-example test set with natural label distribution. For Level-2, the open-weight benchmark used 1,200 balanced training samples and a 400-example balanced validation/evaluation split (100 instances per class). Because a separate held-out Level-2 test set is not yet available, that 400-example split serves as the reported evaluation set in Appendix~\ref{app:open_weight_benchmark}.

\section{Extended Benchmark Results}
\label{app:open_weight_benchmark}

This section reports the broader benchmark that complements the selected main-text tables. In addition to traditional ML baselines and API-based prompting, we evaluated representative open-weight models under supervised fine-tuning: one encoder-only model (BERT-base-Chinese) and two decoder-only models (Qwen2.5-7B and Llama-3-8B). Detailed optimization settings are reported in Appendix~\ref{app:reproducibility}. For Level-1, evaluation uses the held-out 1,901-example test set with its natural class distribution. For Level-2, evaluation uses the 400-example balanced validation split available for this task, because a separate held-out test set is not yet available.

\subsection{Level-1 Full Benchmark (R/W/N)}

Table~\ref{tab:appendix_level1_benchmark} reports the complete Level-1 comparison.

\begin{table*}[t]
\centering
\footnotesize
\setlength{\tabcolsep}{4pt}
\begin{tabular}{llcccccc}
\toprule
\textbf{Model} & \textbf{Method} & \textbf{Kappa} & \textbf{Acc.} & \textbf{R-F1} & \textbf{W-F1} & \textbf{N-F1} & \textbf{M-F1} \\
\midrule
\multicolumn{8}{l}{\textit{Traditional ML}} \\
XGBoost & TF-IDF & 0.153 & 0.487 & 0.233 & 0.538 & 0.488 & 0.420 \\
SVM & TF-IDF & 0.175 & 0.508 & 0.260 & 0.539 & 0.527 & 0.442 \\
\addlinespace
\multicolumn{8}{l}{\textit{API-based baselines}} \\
DeepSeek-V3 & Zero-shot & 0.481 & 0.733 & 0.438 & 0.613 & 0.815 & 0.622 \\
Qwen3-235B & Zero-shot & 0.570 & 0.780 & 0.561 & 0.705 & 0.834 & 0.700 \\
GPT-4o-mini & Zero-shot & 0.618 & 0.793 & 0.482 & 0.721 & 0.873 & 0.692 \\
GPT-4o-mini & Few-shot (k=3) & 0.686 & 0.830 & 0.489 & 0.786 & 0.898 & 0.724 \\
GPT-4o-mini & Few-shot (k=6) & 0.659 & 0.816 & 0.512 & 0.763 & 0.883 & 0.719 \\
GPT-4o-mini & Fine-tuning & 0.841 & 0.910 & 0.677 & 0.905 & 0.952 & 0.844 \\
\addlinespace
\multicolumn{8}{l}{\textit{Open-weight supervised baselines}} \\
Qwen2.5-7B & QLoRA & 0.689 & 0.824 & 0.637 & 0.816 & 0.859 & 0.771 \\
Llama-3-8B & QLoRA & 0.788 & 0.882 & 0.763 & 0.872 & 0.906 & 0.847 \\
BERT-base-Chinese & Full FT & 0.817 & 0.897 & 0.718 & 0.894 & 0.926 & 0.846 \\
\bottomrule
\end{tabular}
\caption{Full Level-1 benchmark on the held-out 1,901-example test set.}
\label{tab:appendix_level1_benchmark}
\end{table*}

Among open-weight models, BERT-base-Chinese and Llama-3-8B deliver the strongest Level-1 results, substantially outperforming zero-shot API baselines and traditional ML. BERT-base-Chinese is especially competitive given its small size and full reproducibility, while GPT-4o-mini fine-tuning still achieves the highest Kappa overall, indicating the closest agreement with expert labels on the natural-distribution test set. Across model families, the minority \textit{R} class remains the main bottleneck.

\subsection{Level-2 Full Benchmark (B/C/G/Y)}

Table~\ref{tab:appendix_level2_benchmark} reports the full Level-2 comparison, and Table~\ref{tab:appendix_level2_recall} reports per-class recall.

\begin{table*}[t]
\centering
\footnotesize
\setlength{\tabcolsep}{4pt}
\begin{tabular}{llccccccc}
\toprule
\textbf{Model} & \textbf{Method} & \textbf{Kappa} & \textbf{Acc.} & \textbf{B-F1} & \textbf{Y-F1} & \textbf{C-F1} & \textbf{G-F1} & \textbf{M-F1} \\
\midrule
Qwen2.5-7B & QLoRA & 0.757 & 0.818 & 0.941 & 0.890 & 0.746 & 0.684 & 0.816 \\
Llama-3-8B & QLoRA & 0.760 & 0.820 & 0.897 & 0.880 & 0.775 & 0.731 & 0.821 \\
BERT-base-Chinese & Full FT & 0.800 & 0.850 & 0.923 & 0.900 & 0.824 & 0.747 & 0.849 \\
GPT-4o-mini & Fine-tuning & 0.833 & 0.875 & -- & -- & -- & -- & -- \\
\bottomrule
\end{tabular}
\caption{Level-2 benchmark on the 400-example balanced evaluation split. GPT-4o-mini source logs report Kappa and Accuracy but not per-class F1.}
\label{tab:appendix_level2_benchmark}
\end{table*}

\begin{table*}[t]
\centering
\footnotesize
\setlength{\tabcolsep}{6pt}
\begin{tabular}{lcccc}
\toprule
\textbf{Model} & \textbf{B Recall} & \textbf{Y Recall} & \textbf{C Recall} & \textbf{G Recall} \\
\midrule
Qwen2.5-7B & 96.0\% & 89.0\% & 78.0\% & 64.0\% \\
Llama-3-8B & 87.0\% & 88.0\% & 81.0\% & 72.0\% \\
BERT-base-Chinese & 90.0\% & 95.0\% & 84.0\% & 71.0\% \\
GPT-4o-mini & 92.0\% & 89.0\% & 85.0\% & 84.0\% \\
\bottomrule
\end{tabular}
\caption{Level-2 per-class recall / accuracy. Because the evaluation split is balanced with 100 examples per class, recall equals per-class accuracy.}
\label{tab:appendix_level2_recall}
\end{table*}

BERT-base-Chinese is the strongest open-weight model on Level-2, outperforming both 7B/8B QLoRA baselines in Kappa, Accuracy, and Macro-F1. Even so, fine-tuned GPT-4o-mini remains best overall. The gap is driven primarily by the \textit{Grey (Ambiguous)} class: GPT-4o-mini reaches 84.0\% recall on \textit{G}, compared with 71.0\% for BERT-base-Chinese and 72.0\% for Llama-3-8B. This difficulty in separating ambiguity from flexible implementation is the main reason we keep GPT-4o-mini as the production model in the main text despite BERT's strong open-weight performance.

\section{Bilingual Annotated Example}

\subsection{R (Red - Prohibitive)}

Chinese: \\
\zh{地方党委和政府要落实属地责任，严格执行国家统一的防控政策，严禁随意封校停课、停工停产、未经批准阻断交通、随意采取``静默''管理、随意封控、长时间不解封、随意停诊等各类层层加码行为。}\\

English:\\
Local Party committees and governments must fulfill territorial responsibilities and strictly implement unified national prevention and control policies. They are strictly prohibited from arbitrarily closing schools, suspending work and production, blocking transportation without approval, imposing so-called ``lockdown'' management at will, or extending restrictions without justification.\\

Rationale:\\
This is labeled as R because it explicitly prohibits specific government actions using strong negative directives such as \zh{严禁}, indicating clear restrictions on administrative behavior.\\

\subsection{B (Black - Authorizing)}

Chinese: \\
\zh{建设城市大数据平台，构建多元异构数据融合的城市运行管理体系，实现对城市基础设施和城市绿地、湿地等重要生态要素的全面感知以及对城市复杂系统运行的深度认知。}\\

English:\\
Build urban big data platforms and establish integrated management systems that combine heterogeneous data sources, enabling comprehensive monitoring of infrastructure and ecological elements, as well as deeper understanding of complex urban systems.\\

Rationale:\\
This is labeled as B because it authorizes and assigns routine administrative tasks to promote smart cities.\\

\subsection{C (Charcoal - Flexible)}

Chinese: \\
\zh{各级人民政府和有关部门应当采取多种形式，培养和训练幼儿园、托儿所的保教人员，推动建立以专职为主、专兼结合、数量充足、结构合理、素质优良的保教人员队伍。}\\

English:\\
Governments at all levels and relevant departments should adopt multiple forms of approaches to train and develop personnel in kindergartens and childcare institutions, and promote the establishment of a workforce that is primarily full-time, complemented by part-time staff, with sufficient numbers, a well-balanced structure, and high professional quality.\\

Rationale:\\
This is labeled as C because it contains ``should'' (\zh{应当}), which signals a directive, while ``multiple forms'' (\zh{多种形式}) grants flexibility in how to achieve it.\\

\subsection{G (Grey - Ambiguous)}

Chinese: \\
\zh{各地在疫情发生后，要及时划定高风险区… 在疫情传播风险不明确或存在广泛社区传播的情况下，可适度扩大高风险区划定范围。}\\

English:\\
After an outbreak occurs, local authorities should promptly designate high-risk areas. When transmission risks are unclear or widespread community transmission exists, the scope of high-risk areas may be moderately expanded.\\

Rationale:\\
This is labeled as G because, although ``should promptly'' (\zh{要及时}) indicates an expected action, the qualifier ``may be moderately'' (\zh{可适度}) introduces ambiguity and discretion.\\

\subsection{Y (Yellow - Pressuring)}
Chinese: \\
\zh{新一代人工智能发展规划是关系全局和长远的前瞻谋划。必须加强组织领导，健全机制，瞄准目标，紧盯任务，以钉钉子的精神切实抓好落实，一张蓝图干到底。}\\

English:\\
The development plan for next-generation artificial intelligence is a forward-looking strategy that bears on the overall situation and long-term development. Organizational leadership must be strengthened, mechanisms must be improved, goals must be clearly targeted, tasks must be closely tracked, and implementation must be carried out in a rigorous and persistent manner, seeing the blueprint through to the end.\\

Rationale:\\
This is labeled as Y because it uses high-pressure and imperative language, including \zh{必须} and \zh{紧盯任务}, signaling strong enforcement, leadership responsibility, and a focus on ensuring full goal completion.\\

\section{Benchmark Error}
\label{app:benchmark_error}
During the first round of coding, our accuracy reached 90.85\% across a total of 1,901 manually labeled paragraphs in the validation set, making 174 errors in total. Our model performs best on R with 95.6\% accuracy, and N category with 92\% accuracy, with W lagging slightly behind at 88.7\% accuracy. The largest error type is misclassifying W as R, which alone accounts for 43.7\% of all mistakes, followed by misclassifying N as W (31.6\% of all mistakes) misclassifying N as R (14.9\%), and W as N (6.9\%). Taken together, the model reflects a tendency to overestimate the number of R and underestimate N. The following are a few representative errors for each error type.\\

\textbf{W $\rightarrow$ R} The model misinterprets statements about government authority or responsibilities as prohibitions against government actions. In simple terms, it mistakes an authorization to punish wrongdoing for a prohibition on state action.
\begin{itemize}
  \item \zh{全国人民代表大会有权罢免国务院组成人员、最高人民法院院长和最高人民检察院检察长。} The National People's Congress has the power to remove members of the State Council, the President of the Supreme People's Court, and the President of the Supreme People's Procuratorate.
  \item \zh{故意毁损人民币的，由公安机关给予警告，并处1万元以下的罚款。} For intentionally damaging RMB, the public security authority shall issue a warning and may impose a fine of up to 10,000 yuan.
\end{itemize}

\textbf{N $\rightarrow$ W} The model mistakes general legal procedures, obligations for private actors, or background descriptions for government directives.
\begin{itemize}
  \item \zh{调解委员会调解劳动争议，应当自当事人申请调解之日起三十日内结束。}A mediation committee shall conclude the mediation of a labor dispute within thirty days from the date the party applies for mediation.
  \item \zh{中国公民因公务出境和中国海员因执行任务出境管理办法，另行制订。} The rules governing the official overseas travel of Chinese citizens and the outbound travel of Chinese seafarers performing duties will be formulated separately.
\end{itemize}

\textbf{N $\rightarrow$ R} The model mistakes prohibitions on behaviors of private actors for prohibitions on government actions. 
\begin{itemize}
  \item \zh{代表机构应当遵守中国法律，不得损害中国国家安全和社会公共利益。}Representative offices (of foreign enterprises) must obey Chinese law and shall not harm national security or the public interest.
  \item \zh{外国教育机构不得在中国境内单独设立学校。} Foreign educational institutions may not independently establish schools in China.
\end{itemize}

During the second round of coding, our accuracy reached 87.5\% across 400 manually labeled paragraphs, with 50 errors in total. The  model classifies both B and Y with 93\% accuracy, but was weaker at 81\% for C and 83\% for G. Misclassifying C as G accounts for 26\% of all mistakes, while misclassifying G as C accounts for 14 percent. Smaller error types include C as Y, G as B, G as Y, each at 10\%. Overall, about 90\% of all misclassifications involve misclassifying or failing to identify either C or G. The main difficulty in the second round lies in cleanly separating these two conceptually adjacent categories.\\

\textbf{C $\rightarrow$ G} errors occur when the model interprets ideas such as ``adapt to local conditions'' or ``based on needs'' as weak or optional rather than as clear but adaptable instructions.
\begin{itemize}
  \item \zh{人民武装警察部队应当根据执行任务的需要，加强教育和训练。} The People's Armed Police shall strengthen education and training according to operational needs.
\end{itemize}

\textbf{G $\rightarrow$ C} errors arise when the model confuses conditional and restricted implementation with flexibility.
\begin{itemize}
  \item \zh{推广农业技术，应当选择有条件的农户……进行示范。}Agricultural technology should be promoted… through demonstration among suitable farmers or regions.
\end{itemize}

\textbf{C $\rightarrow$ Y} errors occur when the model mistakes multi-stage targets for strict mandatory enforcement, reading phased rollout as urgency.
\begin{itemize}
  \item \zh{加强调研，为全面推进农村信用社管理体制改革做好准备。各分支行和农村信用社联合组织要围绕当地农村信用社改革和发展中的突出问题，做一些力所能及的调查。}Strengthen research to prepare for the comprehensive reform of the rural credit cooperative management system. Branch offices and rural credit cooperative federations should conduct investigations within their capacity on prominent issues in the reform and development of local rural credit cooperatives.
\end{itemize}

\textbf{G $\rightarrow$ B} errors occur when conditional or negotiated arrangements are mistaken for neutral administrative procedures.
\begin{itemize}
  \item \zh{选举事宜由省军区与人大常委会协商决定。} Election matters are to be decided through consultation between the military district and the local People's Congress.
\end{itemize}

\textbf{G $\rightarrow$ Y} The model captures strong words without accounting for the discretionary nature of commands.
\begin{itemize}
  \item \zh{国家行政机关对于违反国家纪律的工作人员，在追究纪律责任和给予纪律处分的时候，必须本着严肃和慎重的方针，按照所犯错误的性质和情节的轻重，参照本人平常的表现和对错误的认识程度，分别予以适当的纪律处分或者免予处分。}When state administrative organs investigate disciplinary responsibility and impose disciplinary sanctions on staff members who violate state discipline, they must follow a serious and prudent approach, and, based on the nature of the violation, the severity of the circumstances, the person's usual performance, and their understanding of the wrongdoing, impose an appropriate sanction or exempt them from sanction.
\end{itemize}

\textbf{B $\rightarrow$ G} Routine administrative rules are misread as uncertain and optional.
\begin{itemize}
  \item \zh{财政部应会同有关部门逐步建立和健全统一的利用外资的财务、会计制度和债务报告制度，加强财务管理和财政监督。}The Ministry of Finance, together with relevant departments, shall gradually establish and improve unified financial and accounting systems and debt reporting systems for the use of foreign capital, and strengthen financial management and oversight.
\end{itemize}

\section{Relational Database Schema}
\label{app:db_schema}
This section provides the full field-level schema of the CAPC-CG relational database, supplementing the high-level description in Section~4. The complete schema is reported in Table~\ref{tab:db_schema}. It covers both the \texttt{documents\_{}*} tables, which store document-level metadata, and the \texttt{paragraphs\_{}*} tables, which store paragraph-level text and annotations.
\begin{table*}[t]
\renewcommand{\arraystretch}{1.2}
\centering
\small
\begin{tabular}{@{}p{3.8cm}p{12.4cm}@{}}
\toprule
\textbf{Field} & \textbf{Explanation} \\
\midrule
\multicolumn{2}{@{}l}{\textbf{From \texttt{documents\_*} tables}} \\
\textit{DocumentID} & Primary key for the document tables. A unique integer identifier for each policy document. \\
\textit{Title} & Official title of the policy document. \\
\textit{Validity} & Legal validity status (e.g., effective, repealed). \\
\textit{Content} & Full, raw text of the document. \\
\textit{TextLength} & Total number of characters in the document. \\
\textit{EffectivenessLevel} & Hierarchical level of legal force (e.g., National Law, Administrative Regulation). \\
\textit{IssuingDepartment} & Government body or bodies that issued the document. \\
\textit{DocumentNumber} & Official issuance number of the document. \\
\textit{IssueDate} & Date on which the document was promulgated. \\
\textit{ImplementationDate} & Date on which the document took effect. \\
\midrule
\multicolumn{2}{@{}l}{\textbf{From \texttt{paragraphs\_*} tables}} \\
\textit{ID} & Primary key for the paragraph tables. A unique integer identifier for each paragraph. \\
\textit{DocumentID} & Foreign key linking the paragraph to its parent document. \\
\textit{Paragraph\_content} & Raw text of the specific paragraph. \\
\textit{order} & Sequential position of the paragraph within its parent document. \\
\textit{topics} & Topic labels assigned to the paragraph. \\
\textit{NER} / \textit{NER\_DICT} & Final, cleaned, and validated NER results. \\
\textit{Label\_L1} & Level-1 label\\
\textit{Label\_L2} & Level-2 label\\
\bottomrule
\end{tabular}
\caption{Schema Description For the Relational Database}
\label{tab:db_schema}
\end{table*}

\section{Ethical Consideration}
\subsection{Annotator Identity and Training} 
The three annotators are co-authors of this paper. All annotators have native-level proficiency in Chinese with more than two-year experience in analyzing Chinese policy documents. Prior to annotation, they completed a structured onboarding and training program exceeding 30 hours, which covered the annotation codebook, illustrative examples and edge cases, calibration exercises, and procedures for recording uncertainty. Iterative deliberation of annotation disparities informed successive refinements of the guidelines to enhance coding consistency without privileging any theoretical position. 

Annotators participated in codebook revision and were apprised of the study's objectives. Their linguistic competence and substantive expertise were considered necessary for reliable interpretation of nuanced policy texts not amenable to crowdsourced annotation. Anonymized annotator identifiers (A1/A2/A3) are retained in the released dataset to facilitate analysis of inter-annotator disagreement.

\subsection{Recruitment and Compensation} 
Annotators were recruited as part-time research assistants under the NSF–funded project and paid \$18 per hour for their work. This rate is above the applicable local minimum wage and was intended to fairly compensate them for the time and expertise required for careful annotation. All activities related to this research adhered to the NSF Responsible and Ethical Conduct of Research (RECR) guidelines.

\section{Topic Modeling Details}

Our topic classification follows a computer-assisted, human-guided workflow. We began by constructing a portfolio of China's 12th to 14th Five-Year Plans as the basis of topic discovery. These documents serve as authoritative policy blueprints for the Chinese state and provide us with an authoritative statement of the central government's policy agenda.

We then conducted unsupervised topic discovery using Latent Dirichlet Allocation (LDA). We estimated a series of LDA models across alternative topic numbers and selected $K = 15$ based on coherence score performance. For stability, we cross-validated the LDA-derived topic structure with BERTopic, which yielded broadly similar results.

Following the computer-assisted, human-guided logic introduced in \citet{king2017computer}, we used expert review to interpret and refine the computationally identified topics. A small group of China policy specialists examined the candidate clusters, reviewed their most representative terms and documents, and mapped each cluster onto a substantive policy domain.

For each paragraph, LDA estimates a posterior probability distribution over topics. We assign each paragraph to the policy domain corresponding to the topic with the highest estimated probability. To assess the accuracy of this procedure, annotators manually labeled a random sample of 200 paragraphs; the automated assignment matched human judgment in 92\% of cases.

\section{Additional Descriptive Statistics}
\label{app:additional_stats}

This section presents additional descriptive statistics for the CAPC-CG dataset. Tables~\ref{tab:doc_length_by_type}--\ref{tab:level2_class_distribution} report document-level counts and lengths, the Level-1 label distribution, and the full five-category distribution of policy directives.

\begin{table*}[t]
\centering
\centering
\footnotesize
\setlength{\tabcolsep}{3.5pt}
\renewcommand{\arraystretch}{0.98}
\begin{tabular}{@{}lrrr@{}}
\toprule
Type & Count & Mean & Median \\
\midrule
National Law & 3,137 & 3175.13 & 794.0 \\
Administrative Regulation & 11,621 & 2468.72 & 1477.0 \\
Ministerial Rule & 322,280 & 1614.91 & 588.0 \\
\bottomrule
\end{tabular}
\caption{Statistics by Document Type}
\label{tab:doc_length_by_type}
\end{table*}

\vspace{-0.4em}

\begin{table*}[t]
\centering
\centering
\footnotesize
\setlength{\tabcolsep}{3.5pt}
\renewcommand{\arraystretch}{0.98}
\begin{tabular}{@{}lrr@{}}
\toprule
Label & Count & Percentage \\
\midrule
N (Neutral) & 1,472,697 & 55.6\% \\
W (Affirmative) & 1,051,619 & 39.7\% \\
R (Prohibitive) & 123,315 & 4.7\% \\
\bottomrule
\end{tabular}
\caption{Distribution of the Level-1 Categories}
\label{tab:level1_class_distribution}
\end{table*}

\vspace{-0.4em}

\begin{table*}[t]
\centering
\centering
\footnotesize
\setlength{\tabcolsep}{3.5pt}
\renewcommand{\arraystretch}{0.98}
\begin{tabular}{@{}lrr@{}}
\toprule
Label & Count & Percentage \\
\midrule
B (Authorizing) & 410,704 & 34.96\% \\
Y (Pressuring) & 274,683 & 23.38\% \\
C (Flexible) & 224,222 & 19.08\% \\
G (Ambiguous) & 141,996 & 12.09\% \\
R (Prohibitive) & 123,315 & 10.50\% \\
\bottomrule
\end{tabular}
\caption{Distribution of the Five-Category Taxonomy of Policy Directives}
\label{tab:level2_class_distribution}
\end{table*}

\end{document}